\documentclass[sigconf]{acmart}

\makeatletter
\let\@authorsaddresses\@empty

\makeatother
\renewcommand\footnotetextcopyrightpermission[1]{} 
\usepackage{multirow}
\usepackage{makecell}
\usepackage{booktabs}
\usepackage{array} 
\usepackage{graphicx}
\usepackage{rotating}

\usepackage[caption=false,font=footnotesize,labelfont=rm,textfont=rm]{subfig}
\DeclareSubrefFormat{parens}{#1(#2)}
\usepackage[section]{placeins}

\begin{document}

\title{Cross-Domain Pre-training with Language Models for Transferable Time Series Representations}


\author{Mingyue Cheng}
\affiliation{%
  \institution{State Key Laboratory of Cognitive Intelligence, University of Science and Technology of China}
  \city{Hefei, Anhui Province}
  \country{China}}
\email{mycheng@ustc.edu.cn}

\author{Xiaoyu Tao}
\affiliation{%
\institution{State Key Laboratory of Cognitive Intelligence, University
 of Science and Technology of China}
  \city{Hefei, Anhui Province}
  \country{China}}
\email{txytiny@mail.ustc.edu.cn}

\author{Qi Liu*}
\affiliation{%
\institution{State Key Laboratory of Cognitive Intelligence, University
 of Science and Technology of China}
  \city{Hefei, Anhui Province}
  \country{China}}
\email{qiliuql@ustc.edu.cn}

\author{Hao Zhang}
\affiliation{%
\institution{State Key Laboratory of Cognitive Intelligence, University
 of Science and Technology of China}
  \city{Hefei, Anhui Province}
  \country{China}}
\email{zh2001@mail.ustc.edu.cn}

\author{Yiheng Chen}
\affiliation{%
\institution{State Key Laboratory of Cognitive Intelligence, University
 of Science and Technology of China}
  \city{Hefei, Anhui Province}
  \country{China}}
\email{cyh_20@mail.ustc.edu.cn}

\author{Defu Lian}
\affiliation{%
\institution{State Key Laboratory of Cognitive Intelligence, University
 of Science and Technology of China}
  \city{Hefei, Anhui Province}
  \country{China}}
\email{liandefu@ustc.edu.cn}






\begin{abstract}
    Pre-training universal models across multiple domains to enhance downstream tasks is a prevalent learning paradigm. However, there has been minimal progress in pre-training transferable models across domains for time series representation. This dilemma is incurred by two key factors: the limited availability of training set within each domain and the substantial differences in data characteristics between domains. To address these challenges, we present a novel framework, namely CrossTimeNet, designed to perform cross-domain self-supervised pre-training to benefit target tasks. Specifically, to address the issue of data scarcity, we utilize a pre-trained language model as the backbone network to effectively capture the sequence dependencies of the input sequence. Meanwhile, we adopt the recovery of corrupted inputs as a self-supervised optimization objective, taking into account the locality of time series. To address discrepancies in data characteristics, we introduce a novel tokenization module that converts continuous time series inputs into discrete token sequences using vector quantization techniques. This approach facilitates the learning of transferable time series models across different domains. Extensive experimental results on diverse time series tasks, including classification and forecasting, demonstrate the effectiveness of our approach.  Our codes are publicly available\footnote{https://github.com/Mingyue-Cheng/CrossTimeNet}.

\end{abstract}

\keywords{Cross Domain Transferring, Self-Supervised Learning, Time Series Analysis}

\maketitle

\section{Introduction}

Time series analysis~\cite{wu2021autoformer,shifaz2020ts} is a critical challenge in data science, with a wide range of applications that include healthcare diagnostics (e.g., physiological signals) and industrial monitoring (e.g., sensors and the Internet of Things). Recent  works~\cite{zheng2014time,liu2021gated,liu2024adaptive} show that deep learning methods have significantly resolved this area, due to their offering unparalleled scalability and the ability to model complex, nonlinear relationships within domain-specific data compared to classical models. 

However,  directly applying deep learning-based models does not always yield satisfactory results in real-world applications  \cite{tipirneni2022self}. The primary issue is that most advanced network architectures, such as Transformer-based models~\cite{vaswani2017attention}, are data-intensive, necessitating the collection of extensive labeled training data from the specific application scenario. To overcome this limitation, a significant amount of recent research proposes utilizing self-supervised learning (SSL) on vast amounts of unlabeled time series data. The core idea is to leverage the knowledge gained during the self-supervised pre-training phase, thereby minimizing the requirement for extensive training resources.  The main reason behind this phenomenon is that pre-trained models can enhance performance by discerning valuable patterns within the data from the extensive pre-training datasets. Therefore, various strategies have been explored to extract general-purpose features from different angles, including methods like contrastive learning algorithms~\cite{yue2022ts2vec} and denoising autoencoder models~\cite{zerveas2021transformer, dong2024simmtm}. Specifically, the contrastive learning approach is commonly used in discriminative pretraining methods, where models are trained to learn representations by working with constructed positive and negative pairs \cite{woo2022cost}.

Although these current methods are effective, there are still two limitations. The first limitation is that the time series field often faces the challenge of insufficient training data compared to other fields, making it difficult for models to achieve ideal results with a limited training set. The second limitation is that these existing methods often assume that self-supervised training is confined within the same domain, overlooking the importance of learning transferable knowledge across domains. Meanwhile, learning universal representations has almost become the default setting. From our perspective, cross-domain self-supervised pre-training naturally brings several benefits. To begin with, cross-domain learning helps discover underlying correlations and patterns, thereby enhancing the model’s feature understanding and improving its predictive capabilities. In addition, by leveraging a large amount of unlabeled data in different fields, cross-domain learning helps to obtain transferable generalized features and effectively solve the data scarcity problem in specific fields. Furthermore, cross-domain pre-training significantly improves the overall generalization ability of the model, enabling it to perform well on a variety of tasks with minimal domain-specific fine-tuning. 

Inspired by the above analysis, we decide to explore a promising but under-studied self-supervised pre-training paradigm to enhance time series representation capabilities. This research is full of challenges. For one thing, to address the data sparsity problem, we need to design encoders that are easy to train to efficiently capture key features and patterns with limited data, thereby improving prediction performance and model transferability. For another, to cope with the data differences between different domains, we need to develop a method to unify instances from various domains, which is crucial for building a cohesive pre-trained model.

In this work, we propose CrossTimeNet, a novel framework specifically designed for self-supervised pre-training of time series data across different domains.
First, current research tends to use convolutional~\cite{yue2022ts2vec} or Transformer-based networks~\cite{dong2024simmtm}, which are usually randomly initialized and thus require a large amount of training data to learn effective features. To address this issue, we adopt a pre-trained language model as the backbone network to effectively capture the sequential dependencies of the input data using existing knowledge and features. Considering the locality of time series data, we use the restoration of the corrupted input as the self-supervised optimization objective. This approach can capture bidirectional representations and thus better utilize the contextual information of sequence data. Second, to unify instances from multiple domains,
 we provide a carefully designed tokenizer that can convert continuous time series into discrete tokens through a reconstruction optimization process. This tokenizer enables each segment of the raw time series to be assigned its own identity code, effectively bridging the gap caused by data discrepancies across domains. Using these strategies, we can develop a general pre-trained time series model that extracts various types of knowledge and patterns from different domains. To evaluate the effectiveness of CrossTimeNet, we conduct comprehensive experiments on time series classification and forecasting tasks using multiple real-world datasets. The experimental results clearly confirm the superiority of our CrossTimeNet from multiple perspectives. We hope that CrossTimeNet can inspire further research to develop general time series representation models.

\section{Related Work}
The related research primarily falls into two categories: (1) self-supervised time series representation and (2) time series analysis.

\subsection{Self-supervised Time Series Representation}
A considerable amount of recent research~\cite{zhang2023self} has been centered around self-supervised learning for time series representation. Upon reviewing and synthesizing the current body of work in this area, self-supervised pre-training efforts for time series data can be broadly categorized into the following approaches: encoder-decoder models, contrastive learning-based techniques, and denoising auto-encoder based. \textbf{Encoder-decoder models}: The primary philosophy behind this category is to leverage an encoder to transform input time series data into a latent representation, which is then reconstructed back to the original input (or some variant of it) by a decoder. This approach encourages the model to capture essential temporal dynamics and dependencies in the data. \textbf{Contrastive learning-based} techniques: This paradigm focuses on learning representations by distinguishing between similar (positive) and dissimilar (negative) pairs of time series segments~\cite{oord2018representation}. Techniques such as TNC~\cite{tonekaboni2021unsupervised}, TS-TCC~\cite{eldele2021time}, and TS2Vec~\cite{yue2022ts2vec} fall under this umbrella, each employing unique mechanisms to define and utilize positive and negative samples for training robust time series representations. \textbf{Denoising auto-encoder based} approaches: Methods like TST~\cite{zerveas2021transformer} and SimMTM~\cite{dong2024simmtm} adopt a reconstructive strategy, where the model is trained to predict missing parts of the input time series or reconstruct the series from distorted versions. In addition, \cite{garza2023timegpt, woo2024unified, goswami2024moment,cheng2023timemae,liu2024generative}  have begun to explore the potential of training base models for time series tasks, but this area is still in the exploratory stage and has not yet been fully developed.
\subsection{Time Series Analysis }
Time series analysis has gained significant attention in recent years \cite{middlehurst2023bake, shifaz2020ts}, In classification tasks, its methods can be roughly divided into distance-based methods, interval-based techniques, shapelet-based \cite{ye2009time}, and dictionary-based techniques. \textbf{Distance-based} methods, such as dynamic time warping (DTW) combined with the nearest neighbor classifier (NN-DTW) \cite{ding2008querying}, form the foundation of traditional TSC. \textbf{Interval-based} like time series forest (TSF) \cite{deng2013time} extract features from specific time intervals, while \textbf{shapelet-based} methods focus on identifying predictive sub-sequences. \textbf{Dictionary-based} techniques, exemplified by Symbolic Aggregate approXimation (SAX)~\cite{lin2007experiencing}, transform time series into symbolic representations for analysis. Recently, \textbf{deep Learning-based} methods, particularly convolutional neural networks (CNNs)~\cite{zheng2014time,wang2017time,cheng2024convtimenet} and Transformer architectures~\cite{liu2021gated,cheng2023formertime,wen2022transformers}, have been at the forefront of this wave.
Meanwhile, time series forecasting aims to predict future values based on historical data, with applications in finance and supply chain management. Forecasting models range from classic statistical approaches like ARIMA~\cite{box1968some}, which are limited to stationary sequences, to recurrent neural networks (RNNs) such as LSTM~\cite{hochreiter1997long} and GRU~\cite{chung2014empirical}, which face challenges in training efficiency and long-term dependency modeling. Recently, Transformer models have shown promise in this domain, with notable examples including Informer~\cite{zhou2021informer} and Autoformer~\cite{wu2021autoformer}, which enhance long-term dependency management and computational efficiency.Although deep neural networks have powerful nonlinear modeling capabilities in the field of time series analysis and have the advantage of not requiring manual feature engineering - thus helping to learn more complex temporal features, their main disadvantage is their huge demand for data. These models require a large number of labeled training sets, without which they are prone to overfitting.

\section{Preliminaries}
First, we introduce the studied problem, using notation and concepts.  Then, we briefly present the relative models in our work.

\subsection{Cross-domain Self-supervised}
In this work, our primary goal is to create a unified self-supervised pre-training framework that can efficiently handle time series data from various scenarios, each referred to as a ``domain''. These domains encompass distinct characteristics and patterns within the time series data. To tackle the challenge of self-supervised pre-training across multiple domains, we introduce $\mathcal{D} = {\mathcal{D}_1, \mathcal{D}_2, \ldots, \mathcal{D}_N}$, a collection of datasets representing $N$ distinct domains. Each domain dataset $\mathcal{D}_n$ comprises unlabeled time series data ${\mathbf{x}_i^n}$, with $\mathbf{x}_i^n = [x_{i1}^n, x_{i2}^n, \ldots, x_{iT}^n]$ representing the $i$-th time series instance in the $n$-th domain, consisting of $T$ time points. The key challenge lies in leveraging the vast, unlabeled data across these varied domains to train a versatile pre-trained model $\mathcal{P}$. This model is designed to capture universal features and patterns inherent in time series data, going beyond the specific characteristics of individual domains. The self-supervised learning approach enables the model to uncover and utilize the intrinsic structure of the data without depending on explicit class labels, thereby tapping into the unexploited wealth of unlabeled data across different domains.

Once the pre-trained model $\mathcal{P}$ is established, it can be fine-tuned for corresponding downstream tasks, denoted as $\mathcal{T} = {\mathcal{T}_1, \mathcal{T}_2, \ldots, \mathcal{T}_N}$. These tasks may either fall within the same domains as those used in the pre-training phase or extend across different domains, showcasing the adaptability and transferability of the representations learned by the model. The ultimate aim is to employ the pre-trained model $\mathcal{P}$ to boost the performance and generalization capabilities of models $\mathcal{F}_n$ designated for time series analysis tasks within specific domains. Each model $\mathcal{F}_n: \mathbf{x}_i^n \mapsto y_i^n$ is responsible for mapping a time series to its accurate class label within the respective task domain $\mathcal{T}_n$, thereby enhancing the efficacy of time series classification and forecasting across a broad spectrum of domains. 

\subsection{Pre-trained Language Model}
Recently, pre-trained language models (PLM)~\cite{devlin2018bert} have revolutionized the field of natural language processing (NLP) by providing a powerful framework for learning rich linguistic representations from vast amounts of textual data. At their core, PLMs are models trained on a large corpus of text in a self-supervised manner, meaning they learn to predict parts of the text based on other parts, without the need for explicit annotations or labels. This training approach allows them to capture various language phenomena and contextual nuances, making them highly versatile and effective for various NLP tasks. 
The typical text processing pipeline with a PLM begins with tokenization, where the input text is broken down into manageable pieces, often words or subwords, known as tokens. 
These tokens are then fed into a designed neural network, (e.g., Transformers~\cite{vaswani2017attention}), which processes them to extract dynamic contextual features. As a result, PLMs have paved the way for breakthroughs in NLP by enabling models to generalize across tasks with impressive accuracy and minimal task-specific tuning. 
Due to the powerful capacity shown in PLM, it has nearly become a default paradigm in the NLP domain. 
\section{The Proposed CrossTimeNet}

\begin{figure*}[t]
    \centering
    \includegraphics[width=.99\textwidth]{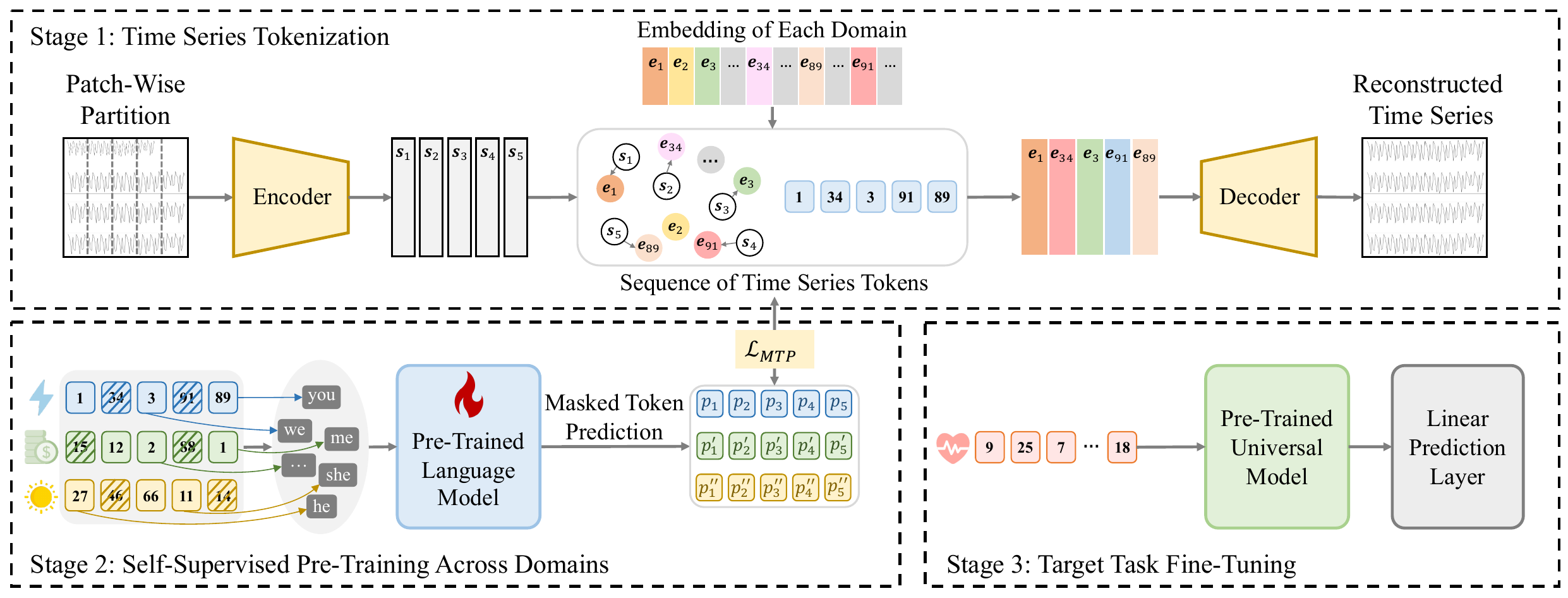}
    \caption{Overview of CrossTimeNet: a unified framework for self-supervised pre-training of cross-domain time series data.}
    \label{fig:framework}
\end{figure*}

\subsection{Overview of Model Architecture}
As depicted in \autoref{fig:framework}, the CrossTimeNet encompasses three core components: (a) time series tokenization, (b) self-supervised pre-training across domains, and (c) domain-specific task fine-tuning. During time series tokenization, a tailor-made tokenizer converts continuous time series data into discrete tokens, establishing a uniform representation suitable for cross-domain applications. Following this, the self-supervised pre-training phase employs a bidirectional token prediction task—where random tokens are masked—to compel the model to deduce missing information, thereby learning potent representations of the time series data. Finally, in the downstream task fine-tuning stage, the model undergoes specific adjustments to excel in domain-related tasks, such as classification, harnessing the extensive knowledge gained from the pre-training process. 
This fine-tuning is meticulously carried out to ensure the model's proficiency in specialized tasks while retaining the extensive insights from its earlier cross-domain exposure. 

\subsection{Time Series Tokenization}
Time series data poses distinct challenges for cross-domain analysis due to its inherent variability in structure, unlike more consistent modalities like text and vision. Variations in channel numbers, physical phenomena representation, and temporal resolutions across domains complicate the development of universally applicable pre-trained models. For example, the simplicity of financial time series contrasts sharply with the complexity of multichannel EEG data. Addressing this, one may wonder that channel-independent modeling~\cite{han2023capacity} emerges as a promising approach, treating each channel as separate to bridge differences between datasets. However, this method may neglect crucial inter-channel dependencies, which are especially vital in fields like healthcare, where understanding the relationships between different physiological signals is key. This oversight could result in a less comprehensive interpretation of the data, highlighting the need for models that can balance generalization with the retention of critical inter-channel information.

In this work, we decide to adopt a direct yet under-explored path - time series discretation during pre-training time series representations. The main idea is to transform the time series into a format that maintains the essential inter-channel relationships while still accommodating the diverse characteristics of data across domains. This discretization not only standardizes the input data but also preserves the rich, multi-faceted information that is crucial for the subsequent stages of self-supervised pre-training and downstream task fine-tuning. Through this approach, we endeavor to construct a robust and versatile pre-trained model, one that harnesses the full spectrum of information within multi-domain time series data while overcoming the inherent challenges posed by their variability. Although some previous works have explored the discretization of time series, including SAX~\cite{lin2007experiencing} and SFA~\cite{schafer2012sfa,schafer2015boss}, these methods suffer from a significant drawback: they are computationally expensive. In addition, these methods exhibit significant limitations when applied to self-supervised pre-training of time series data. They struggle to compress complex temporal sequences effectively, often resulting in substantial information loss. Moreover, these techniques typically require manual tuning, which can introduce subjectivity and limit their scalability.

In light of this, inspired by the success of image compression~\cite{van2017neural}, we employ a auto-encoder framework to achieve efficient time series compression, referred to as time series tokenizer. 
The tokenizer involves several steps. First, we transform the time series along the time dimension into sequence patches. Formally, for a given time series instance $\mathbf{x^n} = [x_1^n, x_2^n, \ldots, x_T^n]$ in the $n$-th domain,  it is first segmented into $L$ non-overlap patches ${\mathbf{s}_1, \mathbf{s}_2, \ldots, \mathbf{s}_L}$, where each patch $\mathbf{s}_l$ contains a subset of consecutive time points from the original series. These patches are then fed into the auto-encoder architecture. In the following sections, we omit the mathematical notation for domains for brevity.

In this work, we propose to map each patch $\mathbf{s}_l$ to a latent representation $\mathbf{z}_n$. The TCN architecture~\cite{oord2016wavenet} is chosen for its ability to capture long-range dependencies and for its efficient computation:

\begin{equation}
	\mathbf{z}_l = \text{Encoder}{\text{TCN}}(\mathbf{s}_l).
\end{equation}

Following the encoding, the vector quantization step occurs. Here, each latent representation $\mathbf{z}_l$ is mapped to the nearest vector in a learned codebook $\mathbf{e} = [\mathbf{e}_1, \mathbf{e}_2, \ldots, \mathbf{e}_K]$, where $K$ is the size of the codebook. This mapping is given by:
\begin{equation}
	\mathbf{z}_l^q = \text{argmin }{\mathbf{e}_k \in \mathbf{e}} | \mathbf{z}_l - \mathbf{e}_k |^2.
\end{equation}\noindent where $\mathbf{z}_l^q $ denote the assigned code selected from codebook. The decoder then reconstructs the time series from the quantized vectors, aiming to minimize the reconstruction error and thus preserve as much of the original sequence information as possible:
\begin{equation}
	\hat{\mathbf{x}} = \text{Decoder}_{\text{TCN}}(\mathbf{z}^q).
\end{equation}

By using this auto-encoder framework with TCNs, we can effectively compress the time series into a discrete representation that is amenable to self-supervised learning tasks while preserving the critical temporal dynamics that are essential for downstream applications. Our optimization process aligns with the approach used in the prior work of VQ-VAE~\cite{oord2016wavenet}, where it is particularly noteworthy that the nearest-neighbor selection process is non-differentiable, leading to challenges in gradient derivation. Due to space constraints in this paper, we omit a detailed discussion, and interested readers may refer to the relevant literature for further information. The novel contribution of our work is that a novel time series discretization is implemented, which is greatly different from previous works.

\subsection{Self-supervised Pre-training}

\subsubsection{Pre-trained Language Model  as Encoders}
\label{sec:encoder_network}
A comprehensive review of the extant literature reveals that the backbone networks in the domain of time series self-supervised learning predominantly harness either convolutional neural networks (e.g., TCN in~\cite{oord2016wavenet} or Transformer networks in~\cite{vaswani2017attention}). In view of the sparsity of data in the time series field, the selected backbone architecture must be easy to train in order to efficiently utilize limited training data. Drawing inspirations from cognitive science literature~\cite{dehaene2017human}, which posits that human learning is not initiated from a tabula rasa but rather resembles a form of pre-trained network, we ventured to adopt a similar paradigm for our backbone network. This approach aligns with the concept that infants' brains, though not fully developed, are equipped with a rudimentary but potent learning framework right from birth. In this vein, our experimental forays led us to an intriguing discovery: employing a pre-trained language model as the backbone network significantly amplifies performance, marking a substantial leap in achieving superior outcomes in time series analysis. This finding is particularly noteworthy as it challenges the conventional neglect of such a potentially efficacious setting in prior research.

Specifically, we draw upon the network architecture of BERT~\cite{devlin2018bert}, renowned for its masked language modeling (MLM) and next sentence prediction (NSP) tasks, to serve as the foundational base model for our network initialization. This choice is predicated on BERT's proven versatility and robustness in capturing contextual dependencies, making it an ideal candidate for our cross-domain self-supervised learning framework. The adoption of a pre-trained language model as the backbone, therefore, represents a novel and promising avenue in the realm of time series analysis, setting the stage for further exploration and validation in future work.

Upon integrating a language model, we encountered the challenge of encoding time series tokens within the BERT network. Although the codebook embedding from tokenizer could be directly utilized, discrepancies in the embedding size and potential gaps in the representational space compared to BERT's word embeddings were apparent. To address this, we employed a word mapping mechanism~\cite{kao2021bert}, randomly assigning each token a corresponding word selected from BERT's vocabulary. This approach effectively resolved the encoding issues of time series tokens, ensuring seamless integration with the language model framework. It should be noted that we provide a detailed comparison of the differences in word mapping to illustrate that the specific impact as shown in \autoref{tab:word_mapping_classification} and \autoref{tab:word_mapping_forecating} presented in Appendix~\ref{sec:word_mapping}.

\subsubsection{Masked Token Prediction}
In this subsection, we introduce the cross-domain self-supervised optimization objectives.  In the realm of time series analysis, in our view, an ideal self-supervised optimization task should fulfill two pivotal objectives. Firstly, it must be capable of learning rich contextual information~\cite{kong2023understanding}, ensuring that the locality of the sequence can be represented (\textit{Objective 1}). This requirement stems from the inherent sequential nature of time series data, where the understanding of a given point is significantly enhanced by its preceding and succeeding elements. Secondly, maintaining a certain level of abstraction of the predicted targets in the self-supervised optimization loss can be beneficial to improving the transferability of pre-trained models (\textit{Objective 2}). This idea is also consistent in \cite{kostelich1993noise}, which thinks that directly formulating self-supervised signals in raw space would largely restrict the capacity of the model due to the noisy and unbound properties. Thirdly, the optimization challenge posed by the self-supervised pre-training should be sufficiently challenge \cite{he2022masked} (\textit{Objective 3}). This difficulty is crucial as it compels the model to learn a more profound and comprehensive set of knowledge, which can be leveraged to augment the performance on downstream tasks. 

Given these considerations, our approach in this paper is to design a self-supervised optimization task characterized by a relatively high ratio (more than 30\%) of \textit{masked token prediction}. This design choice is predicated on the hypothesis that by obscuring a substantial portion of the input data, the model is compelled to infer the masked information based solely on the context provided by the visible data points. Precisely, let $\mathcal{M}$ denotes the set of masked positions within the time series data.  Suppose the predicted corresponding to the masked inputs are $p_l$, with $l\in\mathcal{M}$. Formally, the reconstruction self-supervised optimization goal $\mathcal{L}_{MTP}$ can be described as follows:

\begin{equation}
	\mathcal{L}_{MTP} = -\sum_{l \in \mathcal{M}} \log p(p_l | \mathbf{x}_{\backslash\mathcal{M}}; \Theta),
\end{equation} 
where $\mathbf{x}_{\backslash\mathcal{M}}$ signifies the sequence with the masked tokens removed, and $\Theta$ encapsulates the model parameters. This task not only encourages the model to learn robust representations by leveraging bidirectional context but also ensures that the optimization challenge is sufficiently demanding to facilitate the acquisition of valuable knowledge for downstream applications.

\subsection{Downstream Task Adaptation}
Upon the completion of cross-domain self-supervised pre-training, we have successfully developed a versatile foundation model. To assess the efficacy of the pre-trained model, it is crucial to employ rigorous evaluation techniques. While acknowledging the plethora of transfer learning strategies available, such as Adapter-based methods among others~\cite{hu2021lora,fu2023exploring}, it is pertinent to note that these are beyond the scope of our current investigation and are earmarked for future research endeavors. In this work, we adopt two classical evaluation paradigms within the self-supervised learning framework: \textit{linear evaluation} and \textit{full fine-tuning}.  In linear evaluation, we freeze the weights of the pre-trained model, preserving the learned representations. A linear layer for prediction or classification is then added on top of the model. This layer is the only component that is trained on the downstream task dataset. This approach allows us to evaluate the quality and transferability of the features learned during the self-supervised pre-training phase without modifying the underlying representations. Moreover, linear evaluation is particularly useful for assessing the generalizability of the pre-trained model across different domains with minimal computational cost. Contrary to linear evaluation, full fine-tuning involves adjusting the entire model, including both the pre-trained layers and the newly added task-specific layers. This approach allows the model to fine-tune the learned representations in conjunction with learning the downstream task, potentially leading to higher performance on the target task. Full fine-tuning is more computationally intensive but can result in a model that is more closely tailored to the specifics of the target task, leveraging both the generic representations learned during pre-training and the specific nuances of the new task.

\section{Experiments}
\subsection{Experimental Setup}

To evaluate the effectiveness of our CrossTimeNet, we employ several prevalent real-world datasets representing distinct domains of time series analysis.  For classification tasks, we utilize HAR, ECG, and EEG datasets, while for forecasting tasks, we use ETT-small, Weather, and Exchange datasets. We conduct comparative analyses against a spectrum of prevailing self-supervised baselines, 
 including contrastive learning approaches (TNC~\cite{tonekaboni2021unsupervised}, TS-TCC~\cite{eldele2021time}, TS2Vec~\cite{yue2022ts2vec}), denoising auto-encoder based methods (SimMTM~\cite{dong2024simmtm}, TST~\cite{zerveas2021transformer}). In addition, TST-Zero refers to a Transformer model initialized randomly, while TST-Plus denotes a randomly initialized model that incorporates a pre-trained language model (PLM) architecture. For classification, we apply Accuracy and F1 Score as measurement metrics, while for forecasting, we use Mean Absolute Error (MAE) and Mean Squared Error (MSE). Detailed dataset settings, baseline descriptions, and experimental configurations can be found in Appendix~\ref{sec:data_appendix} and Appendix~\ref{sec:implemental_details_appendix}.

\begin{table*}[t]
  \renewcommand\arraystretch{1.0}
  \setlength{\tabcolsep}{0.08cm}
  \centering
  \caption{Experimental results of time series classification task evaluated by Accuracy and F1 Score.}
  \footnotesize
    \begin{tabular}{c|ccccccccccrrcccc}
    \toprule
    Models & \multicolumn{2}{c}{TNC} & \multicolumn{2}{c}{TS-TCC} & \multicolumn{2}{c}{TS2Vec} & \multicolumn{2}{c}{SimMTM} & \multicolumn{2}{c}{TST} & \multicolumn{2}{c}{TST-Zero} & \multicolumn{2}{c}{TST-Plus} & \multicolumn{2}{c}{CrossTimeNet} \\
    Metric & Accuracy & F1 Score & Accuracy & F1 Score & Accuracy & F1 Score & Accuracy & F1 Score & Accuracy & F1 Score & Accuracy & F1 Score & Accuracy & F1 Score & Accuracy & F1 Score \\
    \midrule
    HAR   & 0.8961  & 0.8951  & 0.8832  & 0.8815  & 0.8968  & 0.8957  & 0.9200 & 0.9220 & 0.9203  & 0.9203  & 0.9121  & 0.9120  & 0.8550  & 0.8520  & \textbf{0.9335 } & \textbf{0.9347 } \\
    EEG   & 0.7603  & 0.4457  & 0.7291  & 0.4347  & 0.7565  & 0.4449  & 0.8165 & 0.6123  & 0.8086  & 0.5516  & 0.7938  & 0.5211  & 0.7929  & 0.5426  & \textbf{0.8541 } & \textbf{0.6402 } \\
    ECG   & 0.2081  & 0.3310  & 0.1178  & 0.3780  & 0.1302  & 0.2064  & 0.2565 & 0.3562  & 0.2206  & 0.3317  & 0.1810  & 0.3861  & 0.2134  & 0.3246  & \textbf{0.4378 } & \textbf{0.6278 } \\
    \bottomrule
    \end{tabular}%
  \label{tab:classification-main}%
\end{table*}%

\begin{table*}[htbp]
  \renewcommand\arraystretch{1.0}
  \setlength{\tabcolsep}{0.18cm}
  \centering
  \caption{Experimental results of time series forecasting task evaluated by MSE and MAE.}
  \footnotesize
    \begin{tabular}{c|cccccccccccccccc}
    \toprule
    Models & \multicolumn{2}{c}{TNC} & \multicolumn{2}{c}{TS-TCC} & \multicolumn{2}{c}{TS2Vec} & \multicolumn{2}{c}{SimMTM} & \multicolumn{2}{c}{TST} & \multicolumn{2}{c}{TST-Zero} & \multicolumn{2}{c}{TST-Plus} & \multicolumn{2}{c}{CrossTimeNet} \\
    Metric & MSE   & MAE   & MSE   & MAE   & MSE   & MAE   & MSE   & MAE   & MSE   & MAE   & MSE   & MAE   & MSE   & MAE   & MSE   & MAE \\
    \midrule
    ETTh1 & 0.6401  & 0.5561  & 0.5962  & 0.5375  & 0.6775  & 0.5690  & 0.5991  & 0.5352  & 0.5718  & 0.5244  & 0.6463  & 0.5571  & 0.5778  & 0.5351  & \textbf{0.5009 } & \textbf{0.4944 } \\
    ETTh2 & 0.4087  & 0.4378  & 0.4122  & 0.4403  & 0.4099  & 0.4395  & 0.4537  & 0.4555  & 0.4171  & 0.4446  & 0.4208  & 0.4496  & 0.4169  & 0.4451  & \textbf{0.3835 } & \textbf{0.4233 } \\
    ETTm1 & 0.6618  & 0.6056  & 0.6323  & 0.5947  & 0.5897  & 0.5765  & 0.4935  & 0.4691  & 0.6030  & 0.5774  & 0.6125  & 0.5869  & 0.5507  & 0.5003  & \textbf{0.4028 } & \textbf{0.4171 } \\
    ETTm2 & 0.3123  & 0.3550  & 0.3225  & 0.3668  & 0.3175  & 0.3611  & 0.3827  & 0.3929  & 0.3113  & 0.3561  & 0.3031  & 0.3514  & 0.3146  & 0.3630  & \textbf{0.2929 } & \textbf{0.3474 } \\
    Weather & 0.3523  & 0.4065  & 0.3023  & 0.3895  & 0.3502  & 0.4172  & 0.3134  & 0.3223  & 0.3067  & 0.3888  & 0.3184  & 0.4035  & 0.3028  & 0.3346  & \textbf{0.2794 } & \textbf{0.3089 } \\
    Exchange & 0.5970  & 0.5606  & 0.6079  & 0.5644  & 0.6540  & 0.5904  & \textbf{0.5684 } & \textbf{0.5345 } & 0.6000  & 0.5605  & 0.6249  & 0.5734  & 0.6540  & 0.5908  & 0.5927  & 0.5499  \\
    \bottomrule
    \end{tabular}%
  \label{tab:forecasting-main}%
\end{table*}%

\subsection{Results and Analysis}
\subsubsection{Downstream  Results}
Based on the experimental results presented in Table~\ref{tab:classification-main} and Table~\ref{tab:forecasting-main}, we can observe CrossTimeNet's performance across various time series tasks, including forecasting and classification. The findings from these tables demonstrate the model's capabilities in learning transferable time series representations and its effectiveness across diverse domains. Due to space limitations, the complete results including linear evaluation are presented in the Appendix \ref{sec:full_result}.

\textbf{Time Series Classification.}
Table~\ref{tab:classification-main} shows that CrossTimeNet achieved the highest scores on all data sets. A key point worth paying attention to is the significant performance differences between different models, especially on ECG datasets. While CrossTimeNet achieved an accuracy of 0.4378 and an F1 score of 0.6278, other models such as TST-Zero and TNC performed significantly worse. This shows that CrossTimeNet performs better in handling the complexity and variability of ECG signals, which is a challenge for other models. From the perspective of time series locality, CrossTimeNet can effectively capture the correlation between adjacent data points, which is crucial for identifying short-term patterns and trends in time series. The extraction of local features enables the model to pay attention to key local changes when processing complex signals, thereby improving classification performance. Especially in ECG signals, subtle changes in heartbeats often contain important clinical information.
A key observation from the table is the significant performance disparity between the models, particularly in the ECG dataset. While CrossTimeNet excels with an Accuracy of 0.4378 and an F1 Score of 0.6278, other models like TST-Zero and TNC show much lower performance. This suggests that CrossTimeNet is better equipped to manage the complexity and variability inherent in ECG signals, which are challenging for other models.
Additionally, the table highlights the varying effectiveness of different models across datasets. For instance, while TST and SimMTM perform relatively well on the HAR dataset, their performance drops significantly on the EEG and ECG datasets. This variability underscores the importance of model selection based on the specific characteristics of the dataset and task at hand. CrossTimeNet's robust performance across all datasets suggests a strong generalization ability, making it a versatile choice for time-series classification tasks.

\textbf{Time Series Forecasting.}
In our time series forecasting experiments, the lookback window length \(L\) is a crucial parameter that determines the number of past observations used to predict future values. We set \(L\) to 336 across all models and datasets, with horizon lengths \(H\) set to [96, 192, 336, 720]. This configuration allows us to evaluate the model's performance across different forecasting lengths and its ability to generalize across various time scales.
The experimental results reported in Table~\ref{tab:forecasting-main}, are averaged over all prediction horizons. The results show that CrossTimeNet generally achieves lower MSE and MAE in multiple datasets. However, on the Exchange dataset, CrossTimeNet performs worse than SimMTM. Specifically, SimMTM achieves an MSE of 0.5684 and an MAE of 0.5345 on the Exchange dataset, both better than CrossTimeNet’s MSE of 0.5927 and MAE of 0.5499. This suggests that when dealing with complex financial time series such as foreign exchange data, SimMTM may be more effective in capturing patterns and changes in the data and providing more accurate predictions. This result also suggests the differences in the adaptability of different models when dealing with different types of data, emphasizing the importance of choosing the right model for the specific task and dataset. Overall, these results highlight the superior performance of CrossTimeNet in capturing complex time dependencies and providing accurate predictions, demonstrating its strong adaptability and generalization capabilities in different scenarios.

\subsubsection{Study of Using PLM as Encoders}
\begin{figure}[b]
  \centering
  \subfloat[PLM for classification]{
    \label{fig:first}
    \includegraphics[width=0.5\linewidth]{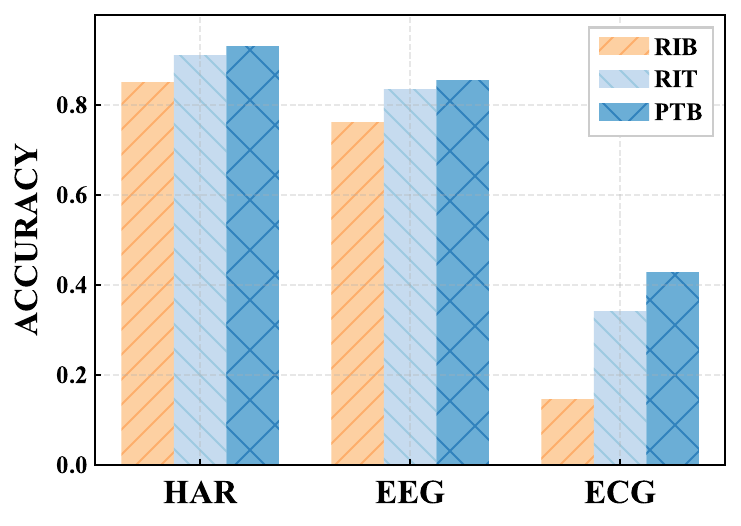}
	}
   \subfloat[PLM for forecasting]{
    \label{fig:second}
    \includegraphics[width=0.5\linewidth]{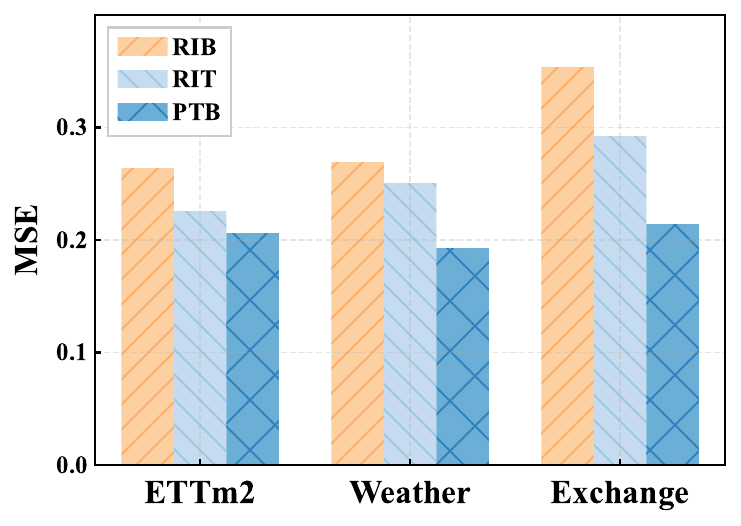}
	}
  \caption{Performance comparison between different PLM}
  \label{fig:PLMs}
\end{figure}

In this study, we evaluate the effectiveness of pre-trained models by comparing the performance of a Pre-trained BERT (PTB), a Randomly Initialized BERT (RIB), and a Randomly Initialized Transformer (RIT) across time series forecasting and classification. Figure \ref{fig:PLMs} displays a performance comparison of these models, highlighting their accuracy in classification and MSE in forecasting tasks.

The exceptional performance of PTB can largely be attributed to its pre-training on a vast corpus of text data, which enables the model to learn complex patterns and dependencies. The rich contextual representations developed during pre-training allow PTB to more effectively parse and predict time series data. Furthermore, leveraging pre-trained knowledge, PTB can adapt more efficiently to specific tasks with limited data, thereby enhancing generalization capabilities and predictive accuracy.

In conclusion, Pre-trained BERT models show significant advantages in handling complex time series tasks. Future research should consider prioritizing pre-trained models to leverage these benefits, as they outperform models without such advanced training, like RIB and RIT, across both classification and forecasting metrics.

\begin{figure}[t]
  \centering
    \subfloat[BERT configurations for classification]{
    \label{fig:first_classification}
    \includegraphics[width=0.49\linewidth]{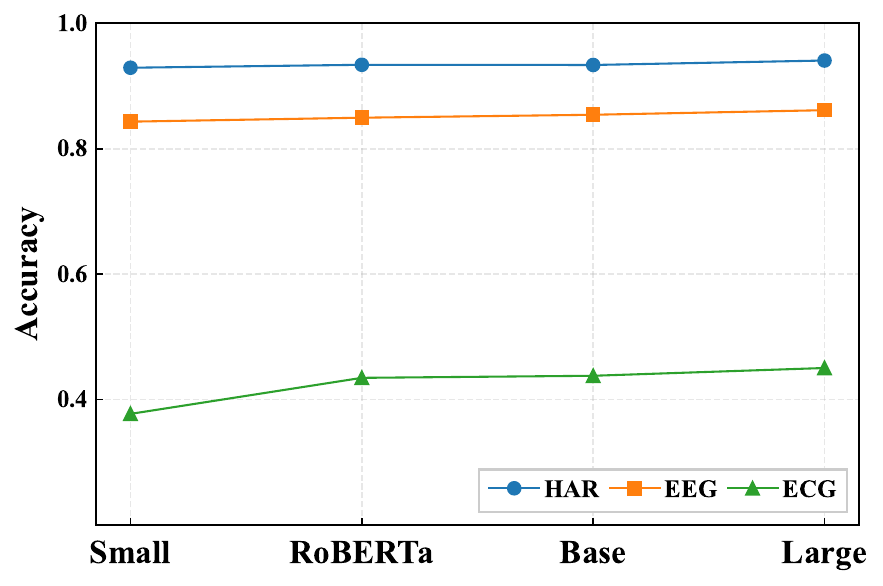}
	}
   \subfloat[BERT configurations for forecasting]{
    \label{fig:second_forecasting}
    \includegraphics[width=0.49\linewidth]{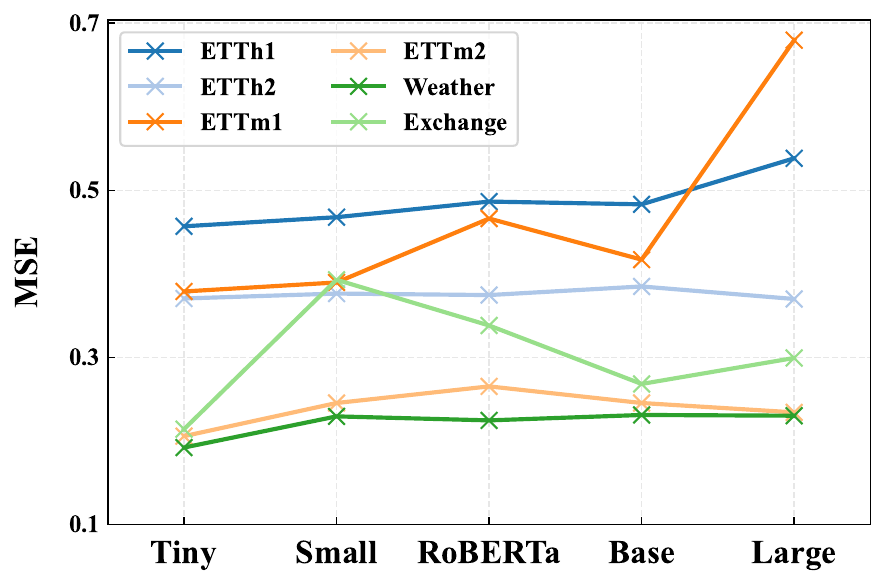}
	}
  \caption{Performance of different BERT  configurations}
  \label{fig:bert_size}
\end{figure}


\subsubsection{Influence of PLM Model Sizes and Structures}
Figure~\ref{fig:bert_size} illustrates the performance of different PLM configurations, comparing BERT-Small to BERT-Large for classification tasks and BERT-Tiny to BERT-Large for forecasting tasks, including RoBERTa across three datasets in CrossTimeNet. Figure \ref{fig:first_classification} shows that for classification tasks, the accuracy is consistently high across all BERT sizes, from BERT-Small to BERT-Large, indicating the minimal impact of model size on performance. This suggests that these tasks have prominent features easily captured by the models.

Figure \ref{fig:second_forecasting} presents the MSE for different BERT sizes across several forecasting tasks, which protection length is 96. Here, BERT-Tiny was included to examine the effect of an extremely small model. Surprisingly, larger models do not significantly reduce MSE and sometimes even increase it, likely due to overfitting, especially with insufficient or noisy data. The inclusion of BERT-Tiny shows that even very small models can perform competitively, indicating that simpler models can be effective and more robust to overfitting in certain cases.

From these findings, larger BERT models generally deliver superior performance in terms of accuracy and F1 score, with BERT-Large being the top performer. This highlights the advantage of extensive model architectures in capturing complex patterns. Inspired by this, scaling the model with mixture-of-expert techniques like Switch Transformer~\cite{fedus2022switch} could be a future direction. RoBERTa, despite its optimizations, does not consistently outperform BERT-Base, possibly due to the specific nature of time-series tasks. Smaller models like BERT-Small and BERT-Tiny still achieve commendable performance, balancing computational efficiency and accuracy. In summary, while model size does not significantly impact classification task performance, larger models may suffer from overfitting in forecasting tasks, leading to higher errors. Therefore, balancing model complexity with task and data characteristics is essential to avoid potential pitfalls of overly complex models.

\begin{table}[b]
  \centering
  \small
  \setlength{\tabcolsep}{0.1cm}
  \caption{Cross-domain pre-training on classification tasks.}
    \begin{tabular}{c|cccccc}
    \toprule
    \multirow{2}[2]{*}{Compared Models} & \multicolumn{2}{c}{HAR} & \multicolumn{2}{c}{EEG} & \multicolumn{2}{c}{ECG} \\
          & Acc & F1 & Acc & F1& Acc & F1  \\
    \midrule
    w/o Cross Domain & 0.9305  & 0.9305  & 0.8541  & 0.6327  & 0.4287  & 0.6161  \\
    w/ Cross Domain & \textbf{0.9335} & \textbf{0.9347} & \textbf{0.8541} & \textbf{0.6402} & \textbf{0.4378} & \textbf{0.6278} \\
    \bottomrule
    \end{tabular}%
  \label{tab:cross_domain_classification}%
\end{table}%

\subsubsection{Impact of Pre-training Across Domains}
The results in Table~\ref{tab:cross_domain_classification} and Table~\ref{tab:cross_domain_forecasting} present the impact of cross-domain integration on both classification and forecasting tasks. 
Table~\ref{tab:cross_domain_classification} shows the classification performance on the HAR, EEG, and ECG datasets, measured by Accuracy and F1 Score. The integration of cross-domain information leads to slight improvements in performance.  The gains are modest across all datasets, indicating that cross-domain integration provides some benefit but is not transformative for classification tasks. This could be because classification tasks often rely heavily on specific features inherent to the dataset, which may not be enhanced by cross-domain data.
Table~\ref{tab:cross_domain_forecasting} illustrates the forecasting performance across different horizons (96, 192, 336, 720) for the ETTh2, ETTm1, and Exchange datasets, evaluated using MSE and MAE. Here, cross-domain integration results in more noticeable improvements, particularly for larger datasets like ETTh2 and ETTm1. For example, in the ETTh2 dataset, the MSE for a 96-point horizon decreases from 0.3706 to 0.3589 with cross-domain integration. However, for the Exchange dataset, the benefits are less pronounced or even detrimental, as seen in the 96-point horizon where MSE increases from 0.2142 to 0.2444. This indicates that cross-domain integration helps capture complex temporal dependencies that are more prevalent in extensive datasets. The mixed results on smaller datasets like Exchange suggest that the additional complexity introduced by cross-domain integration might not always be beneficial, potentially leading to overfitting or increased noise, which can degrade performance. This highlights the importance of tailoring model strategies to the dataset size and characteristics to maximize the benefits of cross-domain integration.

\begin{table}[t]
  \centering
  \caption{Cross-domain pre-training on forecasting tasks.}
    \small  \setlength{\tabcolsep}{0.05cm}
    \begin{tabular}{cc|cccccc}
    \toprule
    \multicolumn{2}{c|}{\multirow{2}{*}{Compared Models}} & \multicolumn{2}{c}{ETTh2} & \multicolumn{2}{c}{ETTm1} & \multicolumn{2}{c}{Exchange} \\
    \multicolumn{2}{c|}{} & MSE   & MAE   & MSE   & MAE   & MSE   & MAE \\
    \midrule
    \multirow{4}[1]{*}{w/o Cross Doamin} & 96    & 0.3706  & 0.4101  & 0.3790  & 0.4105  & 0.2142  & 0.3387  \\
          & 192   & 0.3982  & 0.4302  & 0.4046  & 0.4269  & 0.3329  & 0.4264  \\
          & 336   & 0.4065  & 0.4386  & 0.4523  & 0.4495  & 0.5140  & 0.5336  \\
          & 720   & 0.4400  & 0.4612  & 0.5139  & 0.4799  & 1.2366  & 0.8358  \\ 
    \hline
    \multirow{4}[1]{*}{w/ Cross Domain} & 96    & \textbf{0.3589 } & \textbf{0.4046 } & \textbf{0.3430 } & \textbf{0.3818 } & 0.2444  & 0.3674  \\
          & 192   & \textbf{0.3755 } & \textbf{0.4167 } & \textbf{0.3714 } & \textbf{0.4046 } & \textbf{0.3276 } & \textbf{0.4171 } \\
          & 336   & \textbf{0.3731 } & \textbf{0.4183 } & \textbf{0.4200 } & \textbf{0.4180 } & 0.6046  & 0.5830  \\
          & 720   & \textbf{0.4265 } & \textbf{0.4537 } & \textbf{0.4766 } & \textbf{0.4638 } & \textbf{1.1940 } & \textbf{0.8321 } \\
    \bottomrule
    \end{tabular}%
  \label{tab:cross_domain_forecasting}%
\end{table}%

\begin{table}[b]
  \centering
  \setlength{\tabcolsep}{0.16cm}
  \caption{Performance of model variants using BERT and GPT2 as initialization parameters for the encoder network (Models A and B are two-layer transformers, with A using auto-regression (AR) and B using masking. Models C and D are pre-trained language models (PLMs), with C based on GPT and D on BERT).}
  \small
    \begin{tabular}{c|cccccc}
    \toprule
    \multirow{2}{*}{Model} & \multicolumn{2}{c}{HAR} & \multicolumn{2}{c}{EEG} & \multicolumn{2}{c}{ECG} \\
                              & Acc & F1 & Acc & F1 & Acc& F1\\
    \midrule
    A & 0.8568  & 0.8534  & 0.8001  & 0.5472  & 0.2604  & 0.3978  \\
    B & 0.7618 & 0.7527 & 0.7959 & 0.5234 & 0.165 & 0.2303 \\
    C & 0.9258  & 0.9261  & 0.8353  & 0.5947  & 0.4101  & 0.5949  \\
    D & \textbf{0.9335 } & \textbf{0.9347 } & \textbf{0.8541 } & \textbf{0.6402 } & \textbf{0.4378 } & \textbf{0.6278 } \\
    \bottomrule
    \end{tabular}%
  \label{tab:bert_gpt}%
\end{table}%

\subsubsection{Effectiveness of Masked-style PLM}
In this part, we aim to evaluate the effectiveness of initializing encoder networks with BERT~\cite{devlin2018bert} and GPT2~\cite{radford2019language} parameters for different self-supervised strategies across three datasets for classification. Table~\ref{tab:bert_gpt} presents the performance of various model variants, including two-layer Transformers with Autoregressive (AR) and Masked Token Prediction (MTP) self-supervised pre-training strategies, as well as models utilizing GPT2 and BERT as encoders.  The reported results reveal a clear trend: models initialized with BERT parameters outperform those with GPT2 across all metrics and tasks, particularly when the entire model is fine-tuned. This suggests that BERT's bidirectional training framework may be more conducive to capturing the nuances of these diverse datasets.  This superiority of BERT-based initialization could be attributed to the inherent design that allows it to better understand and integrate the context from both past and future data points in a time series, which is crucial for tasks like HAR, EEG, and ECG analysis where the significance of a data point often depends on its surrounding values. In contrast, GPT-2's forward-only context capture might limit its ability to fully utilize the available temporal information. In summary, the results highlight the importance of choosing an appropriate pre-trained model for initialization based on the nature of the task and the data. For time-series analysis, where contextual understanding from both directions can be crucial, BERT's bidirectional training framework offers a clear advantage over GPT-2's unidirectional approach.


\begin{table}[b]
  \centering
  \setlength{\tabcolsep}{0.14cm}
  \caption{Classification performance of CrossTimeNet across masking ratios.}
  \small
    \begin{tabular}{ccccccc}
    \toprule
    \multicolumn{1}{c}{\multirow{2}[2]{*}{Masking Ratio}} & \multicolumn{2}{c}{HAR} & \multicolumn{2}{c}{EEG} & \multicolumn{2}{c}{ECG} \\
          & ACC & F1 & ACC & F1 & ACC & F1 \\     \midrule
    0.15  & 0.8914  & 0.8923  & 0.8346  & 0.6029  & 0.2081  & 0.3071  \\
    0.30  & 0.9187  & 0.9191  & 0.8332  & 0.5933  & 0.2222  & 0.3313  \\
    0.45  & \textbf{0.8928 } & \textbf{0.8939 } & \textbf{0.8400 } & \textbf{0.5997} & \textbf{0.2430} & \textbf{0.3610} \\
    0.60  & 0.7944  & 0.7890  & 0.8395  & 0.6118  & 0.1940  & 0.2936  \\
    \bottomrule
    \end{tabular}%
  \label{tab:masking_ratio}%
\end{table}%

\subsubsection{Performance Comparison Across Varying Masking Ratios}
In contrast to the common practice in BERT of using a $15\%$ masking rate~\cite{kenton2019bert}, our CrossTimeNet highlights a higher masking rate. Table~\ref{tab:masking_ratio} showcases the impact of varying masking ratios on the final performance. From the shown results, we find that a general trend is that the model's performance is sensitive to the masking ratio, with an optimal range appearing to be around $0.45$, where the model achieves its peak performance across all datasets in terms of both accuracy and F1 score.  A notable anomaly occurs at a masking ratio of 0.60, where a significant drop in performance is observed across all datasets and evaluation methods, indicating that excessive masking may hinder the model's ability to learn effective representations of the data. Additionally, these findings underscore the nuanced relationship between the masking ratio and model generalization, suggesting that higher masking rates are not always beneficial.  These results suggest that while a certain level of input data masking encourages the model to learn more robust and generalizable features, there is a threshold beyond which further masking becomes detrimental, possibly due to the model receiving insufficient information for effective learning. 

This highlights the importance of selecting an appropriate masking ratio tailored to the characteristics of the data and the specific learning task to optimize model performance.

\begin{figure}[t]
    \centering
    \includegraphics[width=0.5\textwidth]{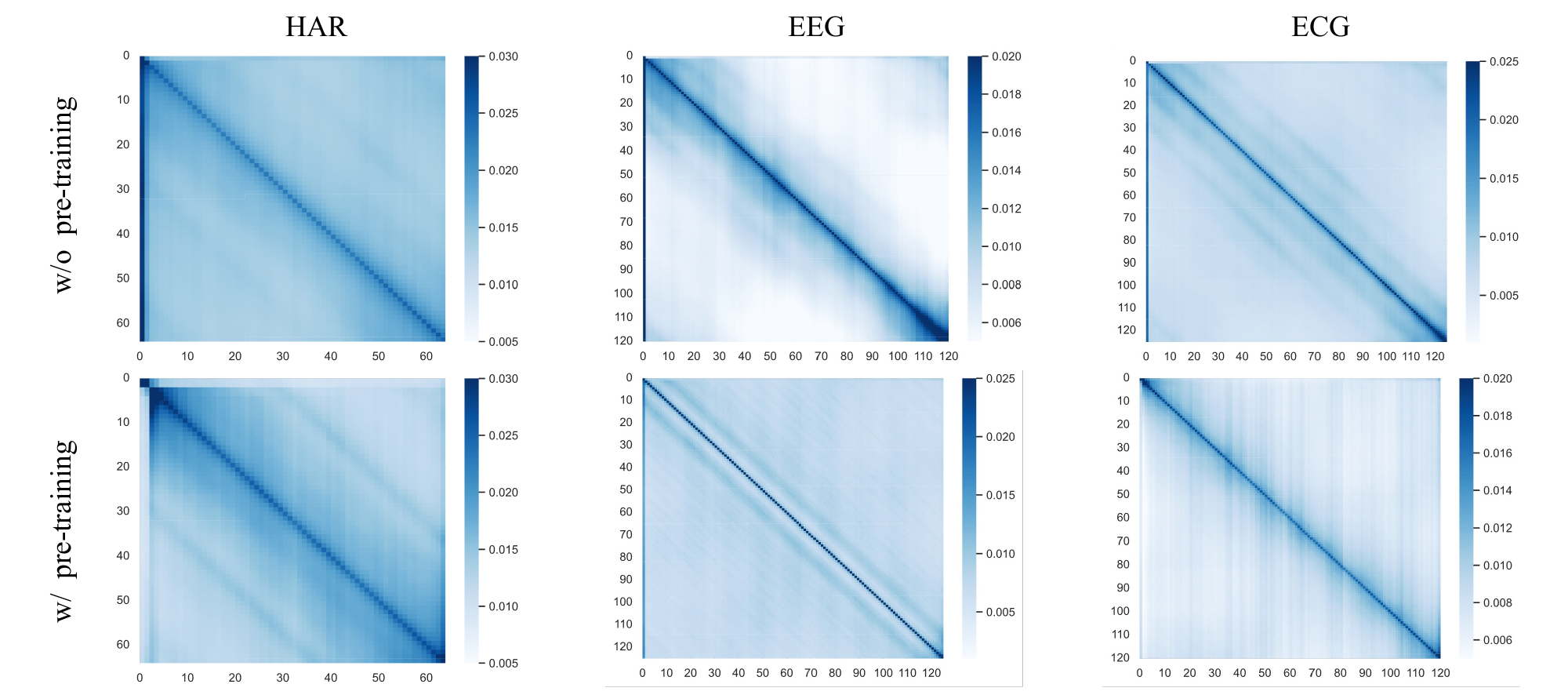}
    \caption{Visualization of attention weight regarding three different datasets.}
    \label{fig:attention_map}
\end{figure}

\begin{figure}[t]
  \centering
    \subfloat[CrossTimeNet]{
    \label{fig:CrossTimeNet}
    \includegraphics[width=0.32\linewidth]{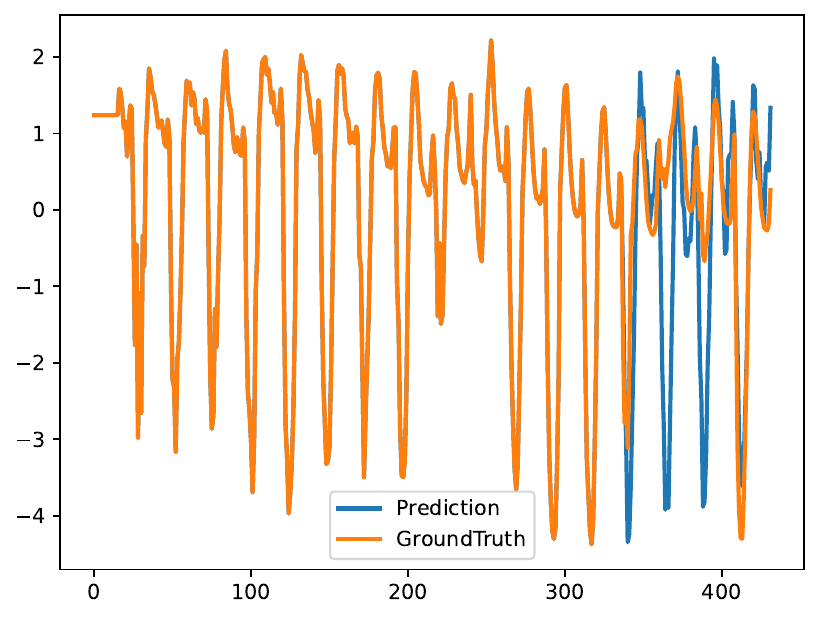}
	}
   \subfloat[TST]{
    \label{fig:TST}
    \includegraphics[width=0.32\linewidth]{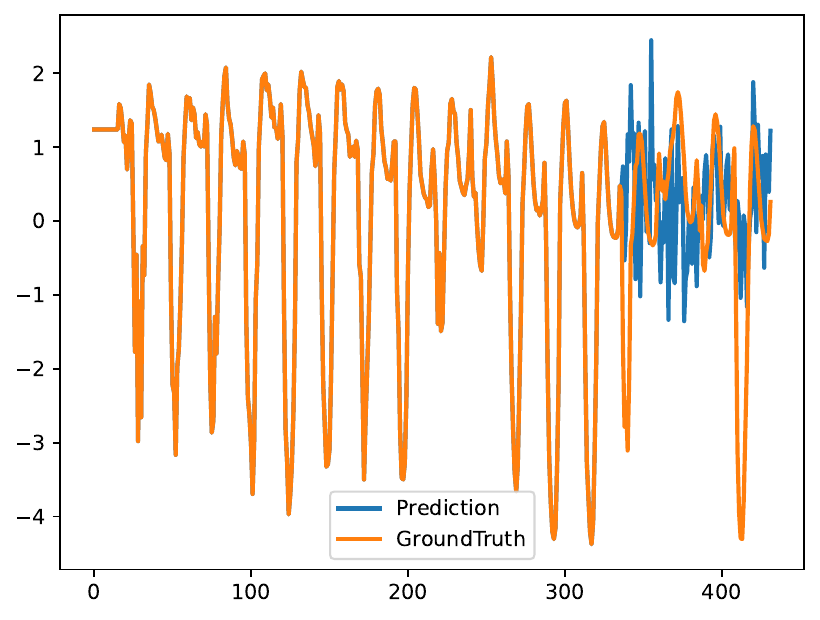}
	}
 \subfloat[Fine-Zero]{
    \label{fig:fine_zero}
    \includegraphics[width=0.32\linewidth]{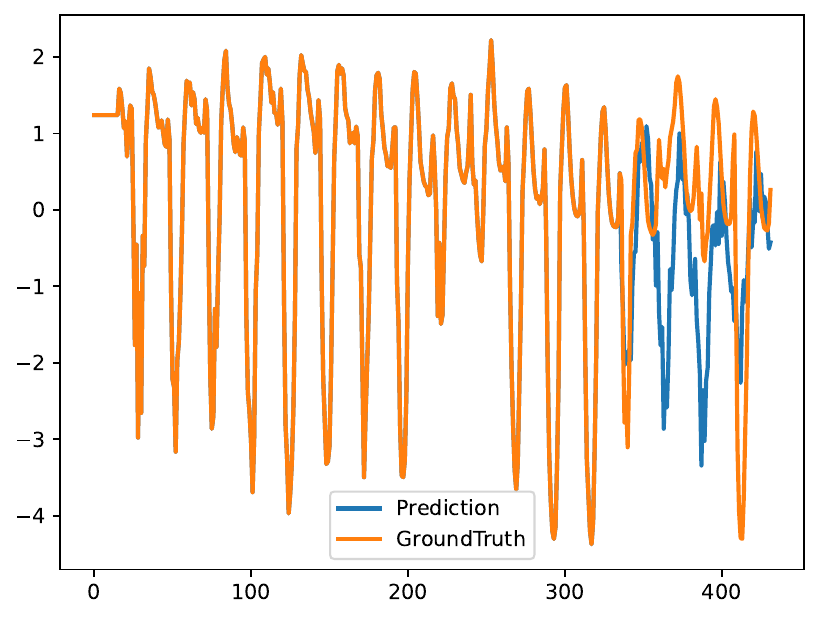}
	}
  \caption{Visualization of forecasting case, generated by various models under the input-336-predict-96 setting, is presented. }
  \label{fig:forecasting_case}
\end{figure}
\subsubsection{Case study analysis}
In the experimental analysis, we investigated the influence of pre-training on attention mechanisms across three distinct datasets: HAR, EEG, and ECG. As shown in  Figure~\ref{fig:attention_map}, we visualized attention weights using heatmaps to discern patterns indicative of the model's focus. For each dataset, we analyzed two conditions: with and without pre-training. The HAR data revealed a pronounced diagonal pattern without pre-training, suggesting a strong self-attention to temporal features, which became more dispersed upon pre-training, indicating a more complex relational understanding. The EEG data presented a less distinct diagonal pattern, with pre-training enhancing temporal structure comprehension. Conversely, the ECG data displayed a consistent focus on temporal autocorrelation, with minimal variations observed between pre-trained and non-pre-trained models. These visualizations underscore the varying impact of pre-training on attention-based models, reflecting the inherent complexities of each dataset.

Figure \ref{fig:forecasting_case}  visualizes forecasting cases generated by various models using 336 input steps and 96 prediction steps. The comparison clearly shows that our approach offers improved accuracy and robustness in capturing complex patterns, particularly in handling irregular fluctuations and intricate temporal dependencies within time series data. This highlights the model's effectiveness in dealing with challenging time series scenarios, validating its superiority.

\section{Conclusion}
\label{sec:conclusion}
In this study, we proposed CrossTimeNet, a novel self-supervised pre-training method designed for time series representation pre-training. Our method's key feature is the time series data discretization, enabling cross-domain self-supervised pre-training. This approach empowers CrossTimeNet to harness temporal dynamics across diverse domains, leading to a versatile and transferable base model. Extensive experiments confirmed CrossTimeNet's effectiveness in learning meaningful and transferable representations, providing substantial benefits for downstream tasks. We hope the CrossTimeNet will inspire more work to be proposed.

\section{Limitation Analysis}
Although the effectiveness of the CrossTimeNet, we also realize that there exist some inherent limitations within our work. This study proposed an innovative self-supervised pre-training method across domains but dose not explore the transferability across different tasks~\cite{cao2023tempo}. Despite showing promising results using PLM as the encoder, the research lacks a comprehensive theoretical explanation for its effectiveness. Additionally, our work has not investigated the potential of generative text models or even large language models for a more universal modeling approach, which could align with the emerging trend of a single model addressing multiple language modeling tasks~\cite{li2023frozen,ouyang2022training}.



\clearpage  
\bibliographystyle{ACM-Reference-Format}
\bibliography{crosstimenet}

\clearpage  
\appendix

\section*{Appendix} 
\section{Detail of Datasets}
\label{sec:data_appendix}
To evaluate the effectiveness of our proposed cross-domain self-supervised pre-training method, we selected three time series datasets for classification and six for forecasting, each drawn from distinct domains. This diverse selection represents a unique set of challenges and characteristics, allowing for a comprehensive assessment of our approach across varied scenarios. The chosen datasets emphasize the adaptability and robustness of our method. 
\subsection{Detail of Classification Datasets}
For the classification task, we selected HAR, EEG, and ECG datasets. The following is a detailed introduction to the datasets.
\begin{itemize}
	\item \textbf{HAR}: The Human Activity Recognition (HAR) dataset is a multi-class classification dataset that consists of sensor data collected from subjects performing various activities. The activities include walking, sitting, standing, and more complex activities like ascending or descending stairs. The data is captured using wearable sensors, providing a rich source of temporal patterns for recognizing human activities.
	\item \textbf{EEG}: The EEG datasets (a.k.a Sleep-EDF) comprise polysomnographic (PSG) recordings, which are used for multi-class sleep stage classification. The dataset includes Electroencephalogram (EEG) recordings among other physiological signals, collected from subjects under normal and pathological conditions. This dataset is pivotal for developing models that can automatically identify sleep stages, aiding in the diagnosis and study of sleep disorders.
	\item \textbf{ECG}: The China Physiological Signal Challenge (CPSC) dataset is a multi-label classification dataset containing Electrocardiogram (ECG) recordings. The dataset is designed for the detection of arrhythmias and other cardiac abnormalities. It provides a diverse set of ECG recordings, making it suitable for developing and evaluating models aimed at cardiac monitoring and diagnosis. 
\end{itemize}
\begin{table}[ht]
	\centering
	\caption{Time series datasets for classification.}
	\vspace{-0.1in}
	\begin{tabular}{cccccc}
		\toprule
		Dataset & \#Train & \#Test & Length & \#Channel & \#Class \\
		\midrule
		HAR   & 8,823  & 2,947  & 128   & 9     & 6 \\
		EEG   & 12,787 & 1,421  & 3,000  & 2     & 8 \\
		ECG   & 10,854 & 1,206  & 5,000  & 12    & 27 \\
		\bottomrule
	\end{tabular}%
	\label{tab:dataset_classification}%
\end{table}%

Table~\ref{tab:dataset_classification} summarizes the key statistics and characteristics of these datasets, including the number of samples, channels, and classes. It is evident that there are significant differences among the datasets in terms of their dimensions and the diversity of classes they encompass. Note that both HAR and EEG data refer to multi-class classification task while ECG involve multi-label classification. It is also worth noting that for the construction of the pre-training dataset, these three datasets were mixed and shuffled together, preserving the train-test splits consistent with previous works in each domain. This approach ensures that our model is exposed to a diverse range of patterns and challenges, simulating real-world scenarios where domain shifts are common.

\subsection{Detail of Forecasting Datasets}
Due to resource constraints, datasets with too many channels, such as the traffic dataset, were not selected. Finally, the ETT, weather, and exchange rates datasets were selected. The following is a detailed introduction to these datasets.

\begin{itemize}
	\item \textbf{ETT}: encompasses records of oil temperature and load metrics of
            electricity transformers. The data spans from July 2016 to
                July 2018, divided into four sub-datasets.
	\item \textbf{Weather}: consists of 21 weather indicators, including air temperature
and humidity. The data was collected at 10-minute intervals throughout the year 2021.
	\item \textbf{Exchange}: contains daily exchange rates of eight different nations. The data spans from 1990 to 2016.
\end{itemize}

\begin{table}[htbp]
  \caption{Time series datasets for forecasting}
  \label{tab:datasets}
  \centering
  \begin{tabular}{ccccc}
    \toprule
    Dataset & Variables & Frequency & Length & Scope \\
    \midrule
    ETTh1/ETTh2 & 7 & 1 Hour & 17420 & Energy \\
    ETTm1/ETTm2 & 7 & 15 Minutes & 69680 & Energy \\
    Exchange & 8 & 1 Day & 7588 & Finance \\
    Weather & 21 & 10 Minutes & 52696 & Weather \\
    \bottomrule
  \end{tabular}
  \label{tab:dataset_forecasting}%
\end{table}

Table ~\ref{tab:dataset_forecasting} presents comprehensive descriptions of the datasets utilized in this study, encompassing a variety of data scenarios and scales. Among these, the ETT dataset primarily focuses on electricity consumption, with four subsets based on varying frequencies. Exchange captures the daily exchange rates among eight countries, while Weather consists of 21 climate indicators, such as air temperature. In accordance with standard protocols, each dataset was split into training, validation, and testing sets based on chronological order. The split ratio for the ETT dataset is 6:2:2, while the Exchange and Weather datasets follow a 7:1:2 ratio. 

\section{Additional Implementation Details}  
\label{sec:implemental_details_appendix}

\subsection{Compared Baselines and Implement Details}
To evaluate the efficacy of our cross-domain self-supervised pre-training approach, we compare it against several recent and popular self-supervised pre-training methods for time series data. These baseline methods encompass both reconstruction-based and contrastive learning approaches, ensuring a comprehensive comparison across different self-supervised learning paradigms.
\begin{itemize}
	
	\item \textbf{TNC }: A contrastive learning approach that treats close temporal segments as positive pairs and distant segments as negative pairs, encouraging the model to learn discriminative features by distinguishing between them.
	\item \textbf{TS-TCC }: This method extends the contrastive learning framework to time series data by leveraging temporal coherence as a signal for similarity, aiming to learn representations that are invariant to specific transformations while maintaining temporal structure.
	\item \textbf{TS2Vec}: A hierarchical contrastive learning approach that captures multi-scale temporal patterns by contrasting representations at different time scales, facilitating a comprehensive understanding of the time series data.
 \item \textbf{SimMTM }: Similar to TST, SimMTM employs a masked autoencoder framework for time series analysis, where portions of the input data are masked, and the model learns to reconstruct the original data. This process enables the model to effectively capture intrinsic temporal dynamics. 
 \item \textbf{TST}: A reconstruction-based self-supervised method uses a Transformer architecture to model time series data by predicting missing segments or forecasting future values. To assess the impact of pre-training, we also consider a TST variant without self-supervised pre-training as a baseline.
 \item \textbf{TST-Zero}: A  method that refers to training vanilla transformer encoder networks from scratch without the use of self-supervised pre-training.
 \item \textbf{TST-Plus}: A  method that integrates the architecture of a pre-trained language model (PLM) but is initialized randomly instead of using pre-trained weights.
	
\end{itemize}

To ensure a fair comparison, all baseline methods employ an encoder network based on the Transformer architecture, maintaining consistency in the model's capacity and structural complexity. During the fine-tuning phase for downstream tasks, the task-specific layer remains identical across all models, ensuring that any observed performance differences can be attributed to the effectiveness of the pre-training strategy rather than variations in the network architecture or task-specific adaptations. In terms of the hyper-parameter settings, the batch size is set to $32$, the remaining ones either strictly follow the specific settings suggested by the original paper or tune the validation sets. We report the results of each baseline under its optimal hyper-parameter settings.  
\subsection{Configurations of Our CrossTimeNet}
Next, we present the details of implementing the CrossTimeNet approach, with a special focus on the time series tokenization process using an autoencoder reconstruction architecture. For the tokenization component, the encoder network is carefully designed with a multi-layer TCN network to ensure efficient encoding of time series data into a compact representation. Specifically, the encoder network contains four layers, which helps to achieve a powerful feature extraction mechanism. The embedding size is set to $64$. For the codebook number, we consistently set it to $512$ in all classification datasets and $256$ in the prediction dataset. As for the patch size, we set it to $2$, $25$, and $40$ in the HAR, EEG, and ECG datasets, respectively, and uniformly set it to $7$ on the prediction task. The network is initialized in a random manner. As for the optimization settings, the learning rate for the classification task is set to $5e^{-4}$, and the learning rate for the prediction task is set to $4e^{-4}$, accompanied by the Adam optimizer. In the pre-training stage, we carefully tuned several hyperparameters to optimize the learning process. These include the learning rate (set to $1e^{-4}$) and the batch size (chosen as $32$). For the fine-tuning phase, we explored two different strategies: full fine-tuning and linear evaluation.
All parameters of the pre-trained CrossTimeNet model are updated during the training of the downstream task while retaining the same hyperparameter settings as the pre-training phase.


\section{Extended Analysis of Experimental Results}  
\label{sec:results_appendix}
\subsection{Results Across Varying Codebook Sizes}
Next, we analyze the performance of CrossTimeNet with varying numbers of tokens in the codebook, as reflected in Table \ref{tab:codebook_size_forecasting} and Table \ref{tab:codebook_size_classification}. The results illustrate how different codebook sizes influence results across different datasets.
The findings indicate a general trend where increasing the number of tokens in the codebook initially enhances both MSE and MAE metrics, particularly evident in the ETTh2 dataset, which shows improved accuracy with a codebook size of 512. However, larger codebook sizes correlate with decreased coverage across all datasets, suggesting potential overfitting. For example, in the Weather dataset, coverage drops from 0.8200 at a codebook size of 128 to 0.7000 at 768, indicating a loss of generalization ability.
In parallel, classification experiments reveal similar trends, where increasing codebook sizes lead to improved accuracy and F1 scores up to a certain threshold, particularly in the HAR dataset, peaking at a codebook size of 768. However, this improvement also comes with increased MSE and decreased coverage, reinforcing the notion of a trade-off between model expressiveness and generalization capability.
In summary, both forecasting and classification results highlight the importance of finding an optimal codebook size that maximizes performance while maintaining efficient data representation, ensuring that CrossTimeNet effectively processes sequential data across various applications.
\begin{table}
	\centering
	\caption{Performance comparison of our CrossTimeNet with different number of token in codebook for classification.}
	\vspace{-0.1in}
	\begin{tabular}{cc|cccc}
		\toprule
		Datasets & Size & Accuracy & F1 Score & Coverage & MSE \\
		\midrule
		\multirow{5}[2]{*}{HAR} & 128   & 0.9298 & 0.931 & 1     & 0.0091 \\
		& 256   & 0.9335 & 0.9348 & 1     & 0.0088 \\
		& 384   & 0.9352 & 0.9366 & 1     & 0.0069 \\
		& 512   & 0.9335 & 0.9347 & 0.7422 & 0.0082 \\
		& 768   & 0.9389 & 0.9401 & 0.6237 & 0.0077 \\
		\midrule
		\multirow{5}[2]{*}{EEG} & 128   & 0.8656 & 0.6549 & 0.9922 & 0.0102 \\
		& 256   & 0.8629 & 0.6335 & 0.4648 & 0.0103 \\
		& 384   & 0.8635 & 0.6596 & 0.4453 & 0.0096 \\
		& 512   & 0.8541 & 0.6402 & 0.2793 & 0.0099 \\
		& 768   & 0.8543 & 0.8384 & 0.1849 & 0.0098 \\
		\midrule
		\multirow{5}[2]{*}{ECG} & 128   & 0.4187 & 0.5995 & 1     & 0.0055 \\
		& 256   & 0.4146 & 0.6054 & 1     & 0.0051 \\
		& 384   & 0.4179 & 0.5953 & 1     & 0.0047 \\
		& 512   & 0.4378 & 0.6278 & 0.5918 & 0.0047 \\
		& 768   & 0.4328 & 0.6118 & 0.4714 & 0.0048 \\
		\bottomrule
	\end{tabular}%
	\label{tab:codebook_size_classification}%
\end{table}%

\begin{table}
	\centering
	\caption{Performance comparison of our CrossTimeNet with different number of tokens in the codebook for forecasting.}
	\vspace{-0.1in}
	\begin{tabular}{cc|cccc}
		\toprule
		Datasets & Size & MSE & MAE & Coverage & MSE \\
		\midrule
		\multirow{5}[2]{*}{ETTh2} & 128   & 0.4501 & 0.4702 & 0.8800 & 0.2036 \\
		& 256   & 0.3706 & 0.4101 & 0.8600 & 0.2036 \\
		& 384   & 0.3908 & 0.4205 & 0.8400 & 0.2100 \\
		& 512   & 0.4100 & 0.4300 & 0.8000 & 0.2150 \\
		& 768   & 0.4305 & 0.4400 & 0.7500 & 0.2200 \\
		\midrule
		\multirow{5}[2]{*}{Weather} & 128   & 0.2100 & 0.2800 & 0.8200 & 0.1900 \\
		& 256   & 0.1922 & 0.2538 & 0.8000 & 0.1850 \\
		& 384   & 0.2000 & 0.2600 & 0.7800 & 0.1800 \\
		& 512   & 0.2150 & 0.2750 & 0.7500 & 0.1750 \\
		& 768   & 0.2300 & 0.2900 & 0.7000 & 0.1700 \\
		\midrule
		\multirow{5}[2]{*}{Exchange} & 128   & 0.2200 & 0.3400 & 0.8400 & 0.2000 \\
		& 256   & 0.2142 & 0.3387 & 0.8100 & 0.1980 \\
		& 384   & 0.2300 & 0.3500 & 0.7900 & 0.1950 \\
		& 512   & 0.2400 & 0.3600 & 0.7500 & 0.1920 \\
		& 768   & 0.2500 & 0.3700 & 0.7000 & 0.1900 \\
		\bottomrule
	\end{tabular}%
	\label{tab:codebook_size_forecasting}%
\end{table}%

\subsection{Results Across Varying Patch Sizes}
The performance of CrossTimeNet with varying patch sizes is summarized in the table\ref{tab:patch_size_forecasting}, which includes metrics such as MSE, MAE, and Coverage across four patch sizes: 7, 14, 21, and 28.
The analysis reveals that a patch size of 7 yields the best overall performance, achieving the lowest MSE (0.3706) and MAE (0.4101) while maintaining a high coverage of 0.8600. This suggests that smaller patch sizes are more effective in capturing the underlying patterns of the data, leading to better model predictions. As the patch size increases to 14 and 21, there is a slight increase in MSE and MAE, with corresponding decreases in coverage, indicating a potential trade-off between model precision and generalization ability.
In parallel, classification experiments in tabel\ref{tab:patch_size_classification} demonstrate similar trends, where smaller patch sizes improve accuracy and F1 scores. However, larger patch sizes can lead to overfitting, as evidenced by increased errors and decreased coverage. This reinforces the notion that while larger patch sizes may seem beneficial for capturing broader patterns, they can hinder the model's ability to generalize effectively.
Overall, both forecasting and classification results highlight the critical impact of patch size on the efficacy of the CrossTimeNet model. Selecting an optimal patch size is essential for maximizing performance, with a patch size of 7 showing the most favorable results in both analyses. This underscores the necessity to balance model granularity with computational efficiency for optimal performance across different tasks.
\begin{table}
	\centering
	\caption{Performance comparison of our CrossTimeNet with different patch sizes for classification.}
	\vspace{-0.1in}
	\begin{tabular}{c|cccc}
		\toprule
		Patch Size & Accuracy & F1 Score & Coverage & MSE \\
		\midrule
		20    & 0.4063  & 0.6051  & 1.0000  & \textbf{0.0033}  \\
		40    & \textbf{0.4287} & \textbf{0.6161} & 0.5918  & 0.0047  \\
		60    & 0.4163  & 0.5935  & 0.4609  & 0.0057  \\
		80    & 0.3856  & 0.5627  & 0.4629  & 0.0066  \\
		100   & 0.3822  & 0.5295  & 0.2891  & 0.0078  \\
		\bottomrule
	\end{tabular}%
	\label{tab:patch_size_classification}%
	\vspace{-0.1in}
\end{table}%

\begin{table}
	\centering
	\caption{Performance comparison of our CrossTimeNet with different patch sizes for forecasting.}
	\vspace{-0.1in}
	\begin{tabular}{c|cccc}
		\toprule
		Patch Size & MSE & MAE & Coverage & MSE \\
		\midrule
		7    &  0.3706&  0.4101& 0.8600 & 0.2063 \\
		14    & 0.3727 & 0.4166 & 0.7780 & 0.2795 \\
		21    & 0.3755 & 0.4176 & 0.6595 & 0.2833 \\
		28    & 0.3873 & 0.4289 & 0.6619 &  0.2846\\
		\bottomrule
	\end{tabular}%
	\label{tab:patch_size_forecasting}%
	\vspace{-0.1in}
\end{table}%
\begin{table*}[htbp]
	\centering
	\caption{Classification performance with varied random word mappings across three repetitions, denoted by A, B, C.}
	\vspace{-0.12in}
	\begin{tabular}{cccccccc}
		\toprule
		\multirow{2}[2]{*}{Evaluation Manners} & \multirow{2}[2]{*}{Models} & \multicolumn{2}{c}{HAR} & \multicolumn{2}{c}{EEG} & \multicolumn{2}{c}{ECG} \\
		&       & Accuracy & F1 Score & Accuracy & F1 Score & Accuracy & F1 Score \\
		\midrule
		\multirow{3}[2]{*}{Full Fine-tuning} & A     & 0.9325  & 0.9339  & 0.8529  & 0.6401  & 0.4395  & 0.6293  \\
		& B     & 0.9345  & 0.9357  & 0.8543  & 0.6403  & 0.4336  & 0.6218  \\
		& C     & 0.9335  & 0.9345  & 0.8550  & 0.6402  & 0.4403  & 0.6324  \\
		\midrule
		\multirow{3}[2]{*}{Linear Evaluation} & A     & 0.9155  & 0.9152  & 0.8388  & 0.5923  & 0.2180  & 0.3259  \\
		& B     & 0.9182  & 0.9189  & 0.8276  & 0.5827  & 0.2231  & 0.3323  \\
		& C     & 0.9223  & 0.9232  & 0.8332  & 0.6050  & 0.2255  & 0.3358  \\
		\bottomrule
	\end{tabular}%
	\label{tab:word_mapping_classification}%
\end{table*}%

\begin{table*}[htbp]
	\centering
	\caption{Forecasting performance with varied random word mappings across three repetitions, denoted by A, B, C.}
	\vspace{-0.12in}
	\begin{tabular}{cccccccc}
		\toprule
		\multirow{2}[2]{*}{Evaluation Manners} & \multirow{2}[2]{*}{Models} & \multicolumn{2}{c}{ETTh2} & \multicolumn{2}{c}{Weather} & \multicolumn{2}{c}{Exchange} \\
		&       & Accuracy & F1 Score & Accuracy & F1 Score & Accuracy & F1 Score \\
		\midrule
		\multirow{3}[2]{*}{Full Fine-tuning} & A     & 0.3706 & 0.4101 & 0.1922 & 0.2538 & 0.2142 &0.3387  \\
		& B  & 0.3710  & 0.4163 & 0.1930 & 0.2503 & 0.2198 & 0.3358   \\
		& C  & 0.3732  & 0.4168 & 0.1998 &0.2535  & 0.2190 &  0.3337  \\
		\midrule
		\multirow{3}[2]{*}{Linear Evaluation} & A     & 0.3906 & 0.4222 & 0.2313 &0.2855  & 0.4467 & 0.4806 \\
		& B     & 0.3947 & 0.4257 &  0.2318& 0.2877 & 0.4472 & 0.4814 \\
		& C     & 0.4175 & 0.4359 &  0.2359& 0.2806 & 0.4571 & 0.4805\\
		\bottomrule
	\end{tabular}%
	\label{tab:word_mapping_forecating}%
\end{table*}%

\subsection{Studying the Token Selection Strategies}
\label{sec:word_mapping}
Table~\ref{tab:word_mapping_classification} and Table \ref{tab:word_mapping_forecating} presents the performance of CrossTimeNet under different random word mappings (denoted as A, B, and C) across three distinct datasets.  Observing the results, it is evident that the variations in word mappings have a relatively minor influence on the overall performance metrics. This result underlines the effectiveness of the word mapping mechanism in bridging the representational gap between time series tokens and the BERT model's vocabulary, thereby enabling the CrossTimeNet to leverage pre-trained language model representations for time series analysis tasks.


\subsection{Data Sparsity Analysis}
\begin{figure*}[htbp]
	\centering
	\includegraphics[width=1\textwidth]{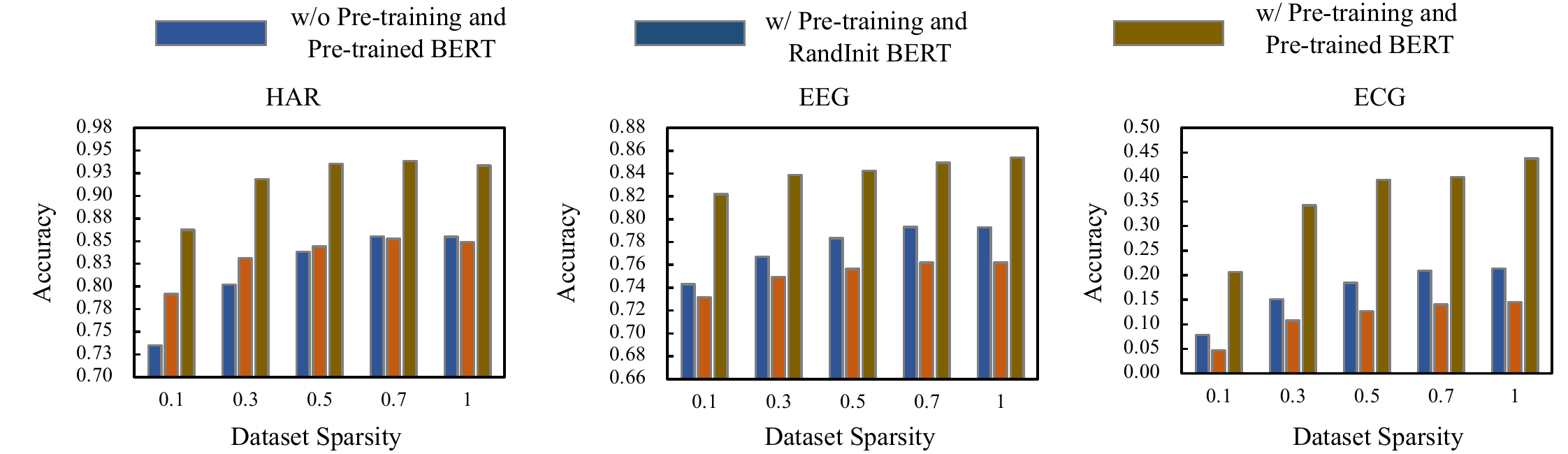}
	\vspace{-0.3in}
	\caption{Investigating the impact of downstream training set sparsity on fine-tuning pre-trained model.}
	\vspace{-0.1in}
	\label{fig:data_sparsity}
\end{figure*}

Figure~\ref{fig:data_sparsity} illustrates the accuracy of fine-tuning a pre-trained model across varying degrees of training set sparsity in three datasets. Overall,  the accuracy of the fine-tuned pre-trained model exhibits a general decline as dataset sparsity increases. This indicates that denser datasets tend to yield better fine-tuning outcomes, affirming the importance of data richness for model performance optimization. Despite the overall trend, the pre-trained BERT model demonstrates a notable degree of robustness to sparsity, maintaining higher accuracy relative to the model without pre-training and the RandomInit BERT. This robustness is especially evident in the ECG domain, where accuracy remains relatively stable until a sparsity level of $0.5$, suggesting that pre-trained models can leverage their learned representations to effectively handle sparse data.  In summary, while dataset sparsity negatively impacts model accuracy, pre-training can serve as a mitigating factor, enhancing the model's ability to maintain performance in sparse data conditions. This underscores the potential of pre-trained models in applications with data limitations.

\subsection{Analyzing Semantic of Time Series Token}
The case study employs t-SNE visualization to articulate the clustering of discrete tokens represented in a high-dimensional feature space, juxtaposed with their manifestation in the corresponding raw time series data. The t-SNE plot distinctly segregates the tokens into coherent clusters, demarcated by a trio of colors, each color signifying a unique token category. This delineation by the t-SNE algorithm evidences its capability to reduce dimensionality while preserving the topology of the dataset, facilitating an intuitive understanding of complex structures within the feature space.

Parallelly, the time series graph delineates these tokens across the temporal axis, with the color-coded segments reflecting the same categorical distinctions identified in the t-SNE visualization. The synchronization between the spatial clusters in the t-SNE plot and the temporal segmentation in the time series data provides a compelling narrative of the data's underlying dynamics. Each color-coded point, representing a discrete token, aligns with a specific behavior or state in the time series, thereby mapping a multidimensional data narrative onto a comprehensible two-dimensional framework.

This analytical approach, combining t-SNE with time series visualization, serves as an effective method for interpreting the nuanced interactions of tokens over time, offering a profound lens through which data scientists can observe temporal patterns, detect anomalies, or even predict future states in sequential data models.

\subsection{Full Results}
\label{sec:full_result}
The comprehensive results of CrossTimeNet on time series forecasting and classification tasks are presented in Tables \ref{tab:classification-main-full} and Table \ref{tab:full_results_forecasting}. In Table \ref{tab:classification-main-full}, the upper portion displays the results for fine-tuning all layers of the model, while the lower portion presents the results for linear evaluation. The 'N/A' placeholder indicates instances where the model's accuracy is effectively zero due to the threshold of measurement precision.
\begin{table*}[hb]
  \centering
  \setlength{\tabcolsep}{0.08cm}
  \caption{Full results of time series forecasting task evaluated by MSE and MAE.}
  \small
    \begin{tabular}{ccc|cccccccccccccccc}
    \toprule
     & \multicolumn{2}{c|}{} & \multicolumn{2}{c}{TNC} & \multicolumn{2}{c}{TS-TCC} & \multicolumn{2}{c}{TS2Vec} & \multicolumn{2}{c}{SimMTM} & \multicolumn{2}{c}{TST} & \multicolumn{2}{c}{TST-Zero} & \multicolumn{2}{c}{TST-Plus} & \multicolumn{2}{c}{CrossTimeNet} \\
          & \multicolumn{2}{c|}{Metric} & MSE   & MAE   & MSE   & MAE   & MSE   & MAE   & MSE   & MAE   & MSE   & MAE   & MSE   & MAE   & MSE   & MAE   & MSE   & MAE \\
    \midrule
    \multicolumn{1}{c}{\multirow{24}[2]{*}{\rotatebox{90}{\parbox{2cm}{Full Fine-tuning}}}} & \multirow{4}[1]{*}{\begin{sideways}ETTh1\end{sideways}} & 96    & 0.6271  & 0.5393  & 0.5817  & 0.5171  & 0.6638  & 0.5498  & 0.5463  & 0.4974  & 0.5587  & 0.5069  & 0.6362  & 0.5396  & 0.5059  & 0.4865  & \textbf{0.4558 } & \textbf{0.4650 } \\
          &       & 192   & 0.6403  & 0.5489  & 0.5942  & 0.5272  & 0.6777  & 0.5601  & 0.5680  & 0.5112  & 0.5746  & 0.5179  & 0.6443  & 0.5478  & 0.5570  & 0.5191  & \textbf{0.4946 } & \textbf{0.4842 } \\
          &       & 336   & 0.6405  & 0.5551  & 0.5961  & 0.5367  & 0.6724  & 0.5654  & 0.5620  & 0.5234  & 0.5736  & 0.5238  & 0.6416  & 0.5539  & 0.5914  & 0.5428  & \textbf{0.5129 } & \textbf{0.4960 } \\
          &       & 720   & 0.6524  & 0.5812  & 0.6129  & 0.5690  & 0.6961  & 0.6008  & 0.7202  & 0.6089  & 0.5804  & 0.5488  & 0.6629  & 0.5869  & 0.6568  & 0.5921  & \textbf{0.5404 } & \textbf{0.5325 } \\
          & \multirow{4}[0]{*}{\begin{sideways}ETTh2\end{sideways}} & 96    & 0.3844  & 0.4175  & 0.3914  & 0.4256  & 0.3892  & 0.4232  & 0.3947  & 0.4204  & 0.3934  & 0.4264  & 0.3874  & 0.4262  & 0.3917  & 0.4250  & \textbf{0.3589 } & \textbf{0.4046 } \\
          &       & 192   & 0.4013  & 0.4318  & 0.4093  & 0.4367  & 0.4068  & 0.4365  & 0.4687  & 0.4544  & 0.4104  & 0.4400  & 0.4094  & 0.4432  & 0.4115  & 0.4421  & \textbf{0.3755 } & \textbf{0.4167 } \\
          &       & 336   & 0.3978  & 0.4351  & 0.4004  & 0.4352  & 0.3996  & 0.4361  & 0.4638  & 0.4628  & 0.4072  & 0.4416  & 0.4121  & 0.4472  & 0.4042  & 0.4405  & \textbf{0.3731 } & \textbf{0.4183 } \\
          &       & 720   & 0.4513  & 0.4668  & 0.4476  & 0.4638  & 0.4441  & 0.4623  & 0.4875  & 0.4843  & 0.4573  & 0.4705  & 0.4743  & 0.4818  & 0.4600  & 0.4728  & \textbf{0.4265 } & \textbf{0.4537 } \\
          & \multirow{4}[0]{*}{\begin{sideways}ETTm1\end{sideways}} & 96    & 0.5710  & 0.5579  & 0.5174  & 0.5332  & 0.4522  & 0.5029  & 0.4607  & 0.4420  & 0.5040  & 0.5191  & 0.5196  & 0.5404  & 0.4918  & 0.4773  & \textbf{0.3430 } & \textbf{0.3818 } \\
          &       & 192   & 0.6361  & 0.5932  & 0.5972  & 0.5853  & 0.5293  & 0.5438  & 0.4691  & 0.4511  & 0.5553  & 0.5498  & 0.5623  & 0.5614  & 0.5451  & 0.4940  & \textbf{0.3714 } & \textbf{0.4046 } \\
          &       & 336   & 0.6866  & 0.6188  & 0.6890  & 0.6186  & 0.6131  & 0.5862  & 0.5024  & 0.4732  & 0.6207  & 0.5893  & 0.6280  & 0.5902  & 0.5660  & 0.5037  & \textbf{0.4200 } & \textbf{0.4180 } \\
          &       & 720   & 0.7536  & 0.6526  & 0.7255  & 0.6416  & 0.7641  & 0.6732  & 0.5417  & 0.5102  & 0.7321  & 0.6513  & 0.7399  & 0.6556  & 0.6000  & 0.5263  & \textbf{0.4766 } & \textbf{0.4638 } \\
          & \multirow{4}[0]{*}{\begin{sideways}ETTm2\end{sideways}} & 96    & 0.2230  & 0.3034  & 0.2566  & 0.3296  & 0.2523  & 0.3235  & 0.2301  & 0.3024  & 0.2358  & 0.3130  & 0.2250  & 0.3058  & 0.2490  & 0.3258  & \textbf{0.2073 } & \textbf{0.2967 } \\
          &       & 192   & 0.2733  & 0.3335  & 0.2905  & 0.3493  & 0.2845  & 0.3427  & 0.3113  & 0.3566  & 0.2766  & 0.3369  & 0.2677  & 0.3313  & 0.2916  & 0.3534  & \textbf{0.2586 } & \textbf{0.3279 } \\
          &       & 336   & 0.3310  & 0.3659  & 0.3304  & 0.3708  & 0.3232  & 0.3640  & 0.4208  & 0.4207  & 0.3229  & 0.3622  & 0.3154  & 0.3581  & \textbf{0.3107 } & \textbf{0.3591 } & 0.3166  & 0.3609  \\
          &       & 720   & 0.4218  & 0.4171  & 0.4125  & 0.4174  & 0.4101  & 0.4142  & 0.5687  & 0.4918  & 0.4100  & 0.4122  & 0.4043  & 0.4103  & 0.4069  & 0.4135  & \textbf{0.3890 } & \textbf{0.4042 } \\
          & \multirow{4}[0]{*}{\begin{sideways}Weather\end{sideways}} & 96    & 0.3743  & 0.3805  & 0.2473  & 0.3578  & 0.2771  & 0.3474  & 0.2054  & 0.2563  & 0.2438  & 0.3450  & 0.2500  & 0.3617  & 0.2506  & 0.3025  & \textbf{0.2044 } & \textbf{0.2581 } \\
          &       & 192   & 0.3066  & 0.3926  & 0.2745  & 0.3739  & 0.3219  & 0.4061  & 0.2645  & 0.2991  & 0.2863  & 0.3767  & 0.3055  & 0.4011  & 0.2802  & 0.3203  & \textbf{0.2563 } & \textbf{0.2915 } \\
          &       & 336   & 0.3385  & 0.4098  & 0.3093  & 0.3879  & 0.3635  & 0.4296  & 0.3702  & 0.3506  & 0.3255  & 0.4016  & 0.3479  & 0.4236  & 0.3145  & 0.3401  & \textbf{0.3057 } & \textbf{0.3300 } \\
          &       & 720   & 0.3899  & 0.4432  & 0.3782  & 0.4383  & 0.4381  & 0.4856  & 0.4136  & 0.3831  & 0.3710  & 0.4317  & 0.3700  & 0.4275  & 0.3658  & 0.3754  & \textbf{0.3513 } & \textbf{0.3561 } \\
          & \multirow{4}[1]{*}{\begin{sideways}Exchange\end{sideways}} & 96    & 0.2466  & 0.3708  & 0.2698  & 0.3859  & 0.3267  & 0.4238  & \textbf{0.2232 } & \textbf{0.3408 } & 0.2648  & 0.3814  & 0.2919  & 0.3997  & 0.3519  & 0.4365  & 0.2444  & 0.3674  \\
          &       & 192   & 0.3735  & 0.4596  & 0.3897  & 0.4653  & 0.4429  & 0.4964  & 0.3438  & 0.4311  & 0.3835  & 0.4619  & 0.4132  & 0.4779  & 0.4575  & 0.5025  & \textbf{0.3276 } & \textbf{0.4171 } \\
          &       & 336   & 0.5570  & 0.5735  & 0.5679  & 0.5703  & 0.6143  & 0.5958  & \textbf{0.5148 } & \textbf{0.5373 } & 0.5575  & 0.5657  & 0.5844  & 0.5779  & 0.6168  & 0.5953  & 0.6046  & 0.5830  \\
          &       & 720   & 1.2109  & 0.8383  & 1.2040  & 0.8360  & 1.2320  & 0.8457  & \textbf{1.1916 } & \textbf{0.8286 } & 1.1942  & 0.8329  & 1.2100  & 0.8382  & 1.1896  & 0.8288  & 1.1940  & 0.8321  \\
    \midrule
    \multicolumn{1}{c}{\multirow{24}[2]{*}{\rotatebox{90}{\parbox{2cm}{Linear Evaluation}}}} & \multirow{4}[1]{*}{\begin{sideways}ETTh1\end{sideways}} & 96    & 0.6281  & 0.5395  & 0.5802  & 0.5168  & 0.6106  & 0.5294  & 0.7100  & 0.5713  & 0.5589  & 0.5076  & 0.6309  & 0.5385  & 0.7139  & 0.5761  & \textbf{0.5230 } & \textbf{0.5018 } \\
          &       & 192   & 0.6410  & 0.5489  & 0.5937  & 0.5272  & 0.6243  & 0.5400  & 0.7200  & 0.5802  & 0.5765  & 0.5189  & 0.6436  & 0.5481  & 0.7225  & 0.5835  & \textbf{0.5513 } & \textbf{0.5150 } \\
          &       & 336   & 0.6400  & 0.5547  & 0.5938  & 0.5359  & 0.6196  & 0.5463  & 0.7176  & 0.5841  & 0.5735  & 0.5339  & 0.6428  & 0.5549  & 0.7206  & 0.5900  & \textbf{0.5852 } & \textbf{0.5389 } \\
          &       & 720   & 0.6508  & 0.5798  & 0.6108  & 0.5679  & 0.6326  & 0.5755  & 0.7357  & 0.6129  & 0.5823  & 0.5496  & 0.6621  & 0.5872  & 0.7446  & 0.6214  & \textbf{0.6353 } & \textbf{0.5836 } \\
          & \multirow{4}[0]{*}{\begin{sideways}ETTh2\end{sideways}} & 96    & 0.3959  & 0.4257  & 0.3900  & 0.4300  & 0.3814  & 0.4196  & 0.3949  & 0.4236  & 0.3871  & 0.4214  & 0.3864  & 0.4255  & 0.3956  & 0.4261  & \textbf{0.3813 } & \textbf{0.4153 } \\
          &       & 192   & 0.4137  & 0.4403  & 0.4086  & 0.4369  & 0.3993  & 0.4334  & 0.4562  & 0.4569  & 0.4035  & 0.4347  & 0.4083  & 0.4424  & 0.4110  & 0.4407  & \textbf{0.4090 } & \textbf{0.4403 } \\
          &       & 336   & 0.4118  & 0.4444  & 0.4010  & 0.4358  & 0.3958  & 0.4343  & 0.4724  & 0.4688  & 0.4004  & 0.4368  & 0.4104  & 0.4463  & 0.4110  & 0.4436  & \textbf{0.4156 } & \textbf{0.4502 } \\
          &       & 720   & 0.4606  & 0.4725  & 0.4460  & 0.4631  & 0.4470  & 0.4646  & 0.5055  & 0.4984  & 0.4502  & 0.4661  & 0.4748  & 0.4821  & 0.4575  & 0.4707  & \textbf{0.4641 } & \textbf{0.4762 } \\
          & \multirow{4}[0]{*}{\begin{sideways}ETTm1\end{sideways}} & 96    & 0.4664  & 0.4545  & 0.5274  & 0.4792  & 0.4470  & 0.4446  & 0.3938  & 0.4146  & 0.4227  & 0.4287  & 0.4192  & 0.4259  & 0.6202  & 0.5217  & \textbf{0.3602 } & \textbf{0.4032 } \\
          &       & 192   & 0.4882  & 0.4665  & 0.5405  & 0.4869  & 0.4676  & 0.4554  & 0.4095  & 0.4241  & 0.4440  & 0.4404  & 0.4401  & 0.4405  & 0.6512  & 0.5368  & \textbf{0.3938 } & \textbf{0.4199 } \\
          &       & 336   & 0.5108  & 0.4792  & 0.5590  & 0.4958  & 0.4902  & 0.4676  & 0.4688  & 0.4591  & 0.4676  & 0.4536  & 0.4639  & 0.4528  & 0.6848  & 0.5503  & \textbf{0.4238 } & \textbf{0.4361 } \\
          &       & 720   & 0.5437  & 0.4986  & 0.5860  & 0.5122  & 0.5227  & 0.4865  & 0.5123  & 0.4885  & 0.5032  & 0.4747  & 0.5023  & 0.4736  & 0.7106  & 0.5641  & \textbf{0.4717 } & \textbf{0.4632 } \\
          & \multirow{4}[0]{*}{\begin{sideways}ETTm2\end{sideways}} & 96    & 0.2401  & 0.3161  & 0.2568  & 0.3296  & 0.2415  & 0.3170  & 0.2258  & 0.3049  & 0.2260  & 0.3057  & 0.2274  & 0.3074  & 0.2624  & 0.3361  & \textbf{0.2192 } & \textbf{0.3006 } \\
          &       & 192   & 0.2859  & 0.3429  & 0.2905  & 0.3492  & 0.2772  & 0.3380  & 0.2827  & 0.3466  & 0.2704  & 0.3319  & 0.2694  & 0.3324  & 0.3050  & 0.3606  & \textbf{0.2644 } & \textbf{0.3305 } \\
          &       & 336   & 0.3379  & 0.3713  & 0.3295  & 0.3700  & 0.3185  & 0.3608  & 0.3521  & 0.3886  & 0.3197  & 0.3592  & 0.3160  & 0.3585  & 0.3398  & 0.3797  & \textbf{0.3055 } & \textbf{0.3533 } \\
          &       & 720   & 0.4264  & 0.4211  & 0.4116  & 0.4169  & 0.4060  & 0.4102  & 0.4820  & 0.4585  & 0.4081  & 0.4102  & 0.4047  & 0.4105  & 0.4220  & 0.4243  & \textbf{0.3913 } & \textbf{0.4027 } \\
          & \multirow{4}[0]{*}{\begin{sideways}Weather\end{sideways}} & 96    & 0.2805  & 0.3746  & 0.2189  & 0.2752  & 0.2100  & 0.2716  & 0.2507  & 0.3059  & 0.2636  & 0.3687  & 0.2343  & 0.2878  & 0.2518  & 0.3034  & \textbf{0.2239 } & \textbf{0.2730 } \\
          &       & 192   & 0.3182  & 0.3927  & 0.2541  & 0.2873  & 0.2449  & 0.2951  & 0.2793  & 0.3234  & 0.3020  & 0.3957  & 0.2725  & 0.3137  & 0.2807  & 0.3212  & \textbf{0.2544 } & \textbf{0.2971 } \\
          &       & 336   & 0.3562  & 0.4200  & 0.2733  & 0.3097  & 0.2851  & 0.3201  & 0.3127  & 0.3433  & 0.3426  & 0.4190  & 0.3116  & 0.3383  & 0.3255  & 0.3498  & \textbf{0.2940 } & \textbf{0.3203 } \\
          &       & 720   & 0.3830  & 0.3845  & 0.3380  & 0.3517  & 0.3406  & 0.3600  & 0.3646  & 0.3764  & 0.3700  & 0.3620  & 0.3691  & 0.3762  & 0.3790  & 0.3842  & \textbf{0.3457 } & \textbf{0.3553 } \\
          & \multirow{4}[1]{*}{\begin{sideways}Exchange\end{sideways}} & 96    & 0.2846  & 0.3973  & 0.2718  & 0.3877  & 0.3063  & 0.4123  & 0.3027  & 0.4128  & \textbf{0.2671 } & \textbf{0.3823 } & 0.3059  & 0.4105  & 0.3380  & 0.4300  & 0.4194  & 0.4830  \\
          &       & 192   & 0.4032  & 0.4765  & 0.3936  & 0.4678  & 0.4214  & 0.4858  & 0.4095  & 0.4811  & \textbf{0.3829 } & \textbf{0.4619 } & 0.4298  & 0.4887  & 0.4470  & 0.4983  & 0.5362  & 0.5470  \\
          &       & 336   & 0.5810  & 0.5870  & 0.5725  & 0.5730  & 0.5949  & 0.5879  & 0.5728  & 0.5779  & \textbf{0.5568 } & \textbf{0.5654 } & 0.6040  & 0.5894  & 0.6121  & 0.5929  & 0.6818  & 0.6309  \\
          &       & 720   & 1.2260  & 0.6430  & 1.2284  & 0.8429  & 1.2199  & 0.8428  & 1.1415  & 0.8132  & \textbf{1.1889 } & \textbf{0.8310 } & 1.2290  & 0.8458  & 1.1700  & 0.8208  & 1.2939  & 0.8658  \\
    \bottomrule
    \end{tabular}%
  \label{tab:full_results_forecasting}%
\end{table*}%

\begin{table*}[b]
  \renewcommand\arraystretch{1.0}
  \setlength{\tabcolsep}{0.08cm}
  \centering
  \caption{Full results of time series classification task evaluated by Accuracy and F1 Score.}
  \vspace{-0.1in}
  \footnotesize
    \begin{tabular}{c|ccccccccccrrcccc}
    \toprule
    Models & \multicolumn{2}{c}{TNC} & \multicolumn{2}{c}{TS-TCC} & \multicolumn{2}{c}{TS2Vec} & \multicolumn{2}{c}{SimMTM} & \multicolumn{2}{c}{TST} & \multicolumn{2}{c}{TST-Zero} & \multicolumn{2}{c}{TST-Plus} & \multicolumn{2}{c}{CrossTimeNet} \\
    Metric & Accuracy & F1 Score & Accuracy & F1 Score & Accuracy & F1 Score & Accuracy & F1 Score & Accuracy & F1 Score & \multicolumn{1}{c}{Accuracy} & \multicolumn{1}{c}{F1 Score} & Accuracy & F1 Score & Accuracy & F1 Score \\
    \midrule
    HAR   & 0.8961  & 0.8951  & 0.8832  & 0.8815  & 0.8968  & 0.8957  & 0.9200 & 0.9220 & 0.9203  & 0.9203  & 0.9121  & 0.9120 & 0.8550  & 0.8520  & \textbf{0.9335 } & \textbf{0.9347 } \\
    EEG   & 0.7603  & 0.4457  & 0.7291  & 0.4347  & 0.7565  & 0.4449  & 0.8165 & 0.6123 & 0.8086  & 0.5516  & 0.7938  & 0.5211 & 0.7929  & 0.5426  & \textbf{0.8541 } & \textbf{0.6402 } \\
    ECG   & 0.2081  & 0.3310  & 0.1178  & 0.3780  & 0.1302  & 0.2064  &0.2565       & 0.3562 & 0.2206  & 0.3317  & 0.1810  & 0.3861  & 0.2134  & 0.3246  & \textbf{0.4378 } & \textbf{0.6278 } \\
    \midrule
    HAR   & 0.8920  & 0.8912  & 0.7713  & 0.7652  & 0.7520  & 0.7504  & 0.7241 & 0.7861 & 0.8337  & 0.8300  & 0.7211  & 0.7120 & 0.7800  & 0.8520  & \textbf{0.9146 } & \textbf{0.9148 } \\
    EEG   & 0.1033  & 0.4643  & 0.7559  & 0.4643  & 0.7347  & 0.4172  & 0.5623 & 0.2281 & 0.6664  & 0.3553  & 0.5538  & 0.2236 & 0.6945  & 0.3762  & \textbf{0.8381 } & \textbf{0.6072 } \\
    ECG   & 0.1051  & 0.0810   & 0.0108  & 0.0140  & N/A  &N/A       & N/A      & N/A  & 0.1033  & 0.0234  & N/A  & N/A  & 0.0531  &0.0818 & \textbf{0.2134 } & \textbf{0.3148 } \\
    \bottomrule
    \end{tabular}%
  \label{tab:classification-main-full}%
\end{table*}%

\begin{figure*}[htbp]
	\centering
	\includegraphics[width=0.7\textwidth]{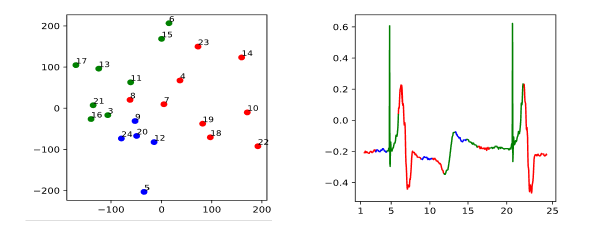}
	\vspace{-0.2in}
	\caption{T-SNE visualization of sequence of patches in a instance and its corresponding raw time series.}
	\label{fig:tsne_sub_series}
\end{figure*}

\end{document}